\documentclass[runningheads]{llncs}
\usepackage{graphicx}
\usepackage{amsmath,amssymb} %
\usepackage{color}

\usepackage{common}
\usepackage{path}

\usepackage{xspace}
\usepackage[bf,small]{caption}
\usepackage{multirow}
\usepackage{dsfont}
\usepackage{subfig}
\usepackage{algorithmic,algorithm}

\newcommand{\comment}[1]{}

\newcommand{\st}{{\rm s.t.}\xspace}
\def\h{{\hbar}}

\def\bh{{\mathds h}}
\def\dh{{\delta}{\mathds  h}}
\def\dhh{{\delta}{h}}
\def\dhat{{\delta}{\bf \hat{\mathds  h}}}

\newcommand{\ones}{{\mathds{1}}}
\def\T{{\!\top}}

\def\w{{\boldsymbol w}}
\def\wh{{\bf \hat{\w}}}
\def\x{{\boldsymbol x}}
\def\bV{{V}}
\def\bmu{{\boldsymbol \mu}}
\def\btheta{{\boldsymbol \theta}}

\def\blambda{{\boldsymbol \lambda}}
\def\balpha{{\boldsymbol \alpha}}

\def\bgamma{{\boldsymbol \gamma}}

\def\Real{\mathbb{R}}
\def\argmax{\operatorname*{argmax\,}}

\def\Scal{\mathcal{S}}

\def\Z{\tau_{max}}
\def\z{\tau}

\def\mb{MultiBoost\xspace}
\def\mbb{MultiBoost\xspace}

\def\mcg{MultiBoost$_{\rm cw}$\xspace}
\def\mcgg{MultiBoost$_{\rm cw}$\xspace}

\def\mcgone{MultiBoost$_{\rm cw}$-1\xspace}

\def\scd{FCD\xspace}

\begin{document}

\newcommand{\point}{
    \raise0.7ex\hbox{.}
    }

\pagestyle{headings}

\mainmatter

\title{Fast Training of Effective Multi-class Boosting Using
       Coordinate Descent Optimization\thanks{
       Code can be downloaded at {\tt http://goo.gl/WluhrQ}.
        This paper was published in Proc.\
        11th Asian Conference on Computer Vision, Korea 2012.
       }
}

\titlerunning{Fast Training of Multi-class Boosting}
\authorrunning{G. Lin, C. Shen, A. van den Hengel, D. Suter} %
\author{Guosheng Lin, Chunhua Shen\thanks{Corresponding author: {\tt chshhen@gmail.com}.},
    Anton van den Hengel, David Suter}
\institute{
The University of Adelaide, Australia
}
\maketitle

\begin{abstract}

    We present a novel column generation based boosting method for
    multi-class classification.
    Our multi-class boosting is formulated in a single optimization
    problem as in \cite{Shen2011Direct,cvpr2012Paisitkriangkrai}.
    Different from most existing multi-class boosting methods, which
    use the same set of weak learners for all the classes,
    we train class specified weak learners (i.e.,
    each class has a different set of weak learners).
    We show that using separate weak learner sets for each class leads
    to fast convergence,
    without introducing additional computational  overhead in the
    training  procedure. To further make the training more efficient
    and scalable,  we also
    propose a fast coordinate descent method for
    solving the optimization problem at each boosting iteration.
    The proposed coordinate descent method is conceptually simple
    and easy to implement in that it is a closed-form solution for
    each coordinate update.
    Experimental results on a variety of  datasets show that,
    compared to a range of existing multi-class boosting methods, the proposed
    method has much faster convergence rate and  better generalization
    performance in most cases. We also empirically show that the
    proposed fast coordinate descent algorithm needs less
    training time than the MultiBoost algorithm in Shen and Hao
    \cite{Shen2011Direct}.

\end{abstract}

\section{Introduction}

    Boosting methods  combine a set of weak classifiers (weak
    learners) to form a strong classifier.
    Boosting has been extensively studied
    \cite{Schapire1998,Shen2011Totally} and
    applied to a wide range of applications due to its robustness and
    efficiency (e.g., real-time object detection
    \cite{Viola2004,AsymBoost2011Wang,Human2008Paul}).
    Despite that fact that most classification tasks are inherently
    multi-class problems, the majority of boosting algorithms are
    designed for binary classification.
    A popular approach to multi-class boosting is to split the
    multi-class problem into a bunch of binary classification
    problems.
    A simple example is the one-vs-all approach. The well-known
    error
    correcting output coding (ECOC) methods \cite{Dietterich1995}
    belong to this category. AdaBoost.ECC \cite{Guruswami1999},
    AdaBoost.MH \cite{Schapire99improvedboosting} and
    AdaBoost.MO \cite{Schapire99improvedboosting} can all be viewed as
    examples of the ECOC
    approach.  The second approach is to
    directly formulate multi-class as a single learning task, which
    is based on pairwise model comparisons between different classes.
    Shen and Hao's  direct
    formulation for multi-class boosting (referred to as \mbb) is such
    an example \cite{Shen2011Direct}. From the perspective of
    optimization,  \mbb\ can
    be seen as an extension of the binary column generation
    boosting framework \cite{Demiriz2002LPBoost,Shen2011Totally}
    to the multi-class case. Our work here builds upon  \mbb.
    As most existing multi-class boosting, for \mbb\ of
    \cite{Shen2011Direct},
    different classes share the same set
    of weak learners, which leads to a sparse solution of the model
    parameters and hence slow convergence. To solve this problem,
    in this work we propose a novel formulation
    (referred to as \mcg) for multi-class boosting by using separate
    weak learner sets. Namely, each class uses its own weak learner set.
    Compared to \mbb, \mcg\ converges much faster, generally has better
    generalization performance and does not introduce additional time cost
    for training.
    Note that AdaBoost.MO proposed in \cite{Schapire99improvedboosting}
    uses different sets of weak classifiers for each class too.
    AdaBoost.MO is based on ECOC and the code matrix in AdaBoost.MO is
    specified before learning. Therefore, the underlying dependence
    between the fixed code matrix and generated binary
    classifiers is not explicitly taken into consideration, compared
    with AdaBoost.ECC.
    In contrast, our \mcg\ is based on the direct formulation of
    multi-class boosting, which leads to fundamentally different
    optimization strategies. More importantly, as shown in our
    experiments, our \mcg\ is much more scalable than AdaBoost.MO
    although both enjoy faster convergence
    than most other multi-class boosting.

    In \mbb \cite{Shen2011Direct}, sophisticated optimization tools like
    Mosek or LBFGS-B \cite{lbfgs} are needed to solve the resulting optimization
    problem at each boosting iteration, which is not very scalable.
    Here we propose a  coordinate descent algorithm (\scd)
    for fast optimization of the resulting problem at each boosting
    iteration of \mcgg. \scd
    methods choose one variable at a time and efficiently solve the
    single-variable sub-problem.
    CD(coordinate decent) has been applied to solve many large-scale optimization problems.
    For example, Yuan et
    al. \cite{Yuan2010JMLR} made comprehensive empirical comparisons of
    $ \ell_1$ regularized classification algorithms.
    They concluded that
    CD methods are very competitive for solving
    large-scale problems.
    In the formulation of \mb (also in our \mcgg), the number
    of variables is the product of the number of classes
    and the number of weak
    learners, which can be very large (especially when the number of
    classes is large). Therefore CD methods may be a  better choice
     for fast optimization of multi-class boosting. Our
    method \scd is specially tailored for the optimization of \mcgg.
    We are able to obtain a closed-form solution for each variable
    update. Thus the optimization can be extremely fast.
    The proposed \scd is easy to implement and no
    sophisticated optimization toolbox is required.

\textbf{Main Contributions}
1) We propose a novel multi-class boosting method (\mcg) that uses class
specified weak learners. Unlike \mb sharing a single set of weak
learners across
different classes, our method uses a separate set of weak learners for each
class. We generate $K$ (the number of classes) weak learners in each
boosting iteration---one weak learner for each class.
    With this mechanism, we are able to achieve much faster
    convergence.
    2) Similar to \mb \cite{Shen2011Direct}, we employ column
    generation to implement the boosting training.
    We derive the Lagrange dual problem of the new multi-class
    boosting formulation which enable us to
    design fully corrective multi-class algorithms using the
    primal-dual optimization technique.
    3) We propose a
    \scd method  for fast training of \mcg. We obtain an
    analytical solution for each variable update in coordinate descent.
    We use the Karush-Kuhn-Tucker (KKT)
    conditions to derive effective stop criteria and
    construct working sets of violated variables for faster
    optimization.
    We show that
    \scd can be applied to fully corrective optimization (updating all
    variables) in multi-class boosting,
    similar to  fast stage-wise optimization in standard AdaBoost
    (updating newly added variables only).

    {\bf Notation}
    Let us assume that we have $ K $ classes.
    A weak learner is a function that maps an example $ \x $ to $\{-1,
    +1 \}$.  We denote each weak learner by $\h$: $ \h_{y,j} ( \cdot,
    \cdot ) \in {\Fcal},(y = 1 \dots K$, and $j = 1 \dots n$).
    ${\Fcal}$
    is the space of all the weak learners; $n$ is the number of weak learners.
        We define column vectors $ \bh_y( \x ) = [ \h_{y,1} ( \x ),
        \cdots, \h_{y,n}(\x) ]^\T$ as the
        outputs of weak learners associated with the $ y $-th class on example $ \x $.
    Let us denote the weak learners' coefficients $\w_y$ for class $ y
    $. Then the strong classifier for class $ y $ is $ F_y  ( \x ) =
    \w^\T \bh_y ( \x )   $.
    We need to learn $ K $ strong classifiers, one for each class.
    Given a test data $ \x$, the classification rule is
    $  y^\star = \argmax F_y( \x) $.
    $\ones$ is a vector with elements all being one. Its dimension
    should be clear from the context.

\section{Our Approach} \label{sec:mcg}
We show how to formulate the multi-class boosting problem in the large
margin learning framework.
Analogue to \mbb, we can define the multi-class margins associate with
training data $ (\x_i, y_i ) $ as
\begin{equation}
\gamma_{(i,y)} = \w_{y_i}^\T  \bh_{y_i}( \x_i) - \w_y^\T \bh_y (\x_i
),
\label{EQ:CS1}
\end{equation} for $ y \neq y_i$.
    Intuitively, $ \gamma_{(i,y)} $ is the difference of the
    classification scores between a ``wrong'' model and the right
    model. We want to make this margin as large as possible.
    \mcg with the exponential loss can
be formulated as:
        \begin{align}
          \min_{ {\w \geq 0, \bgamma}}  %
  \|\w \|_1 +  {\frac{C}{p}} \, \sum_{i} \sum_{y\neq y_i}
  \exp(-\gamma_{(i,y)}),
\forall i=1 \cdots m;
    \forall y \in  \{1\cdots K\} \backslash y_i.
   \label{eq:mcgcon}
\end{align}
Here $ \gamma $ is defined in \eqref{EQ:CS1}. We have also introduced
a shorthand symbol $ p = m \times (K-1)$.
The parameter $ {C} $ controls the complexity of the learned
model.

The model parameter is $ \w = [ \w_1; \w_2; \dots,     \w_K
]^\T \in \Real^{K\cdot n \times 1  } $.

Minimizing \eqref{eq:mcgcon} encourages  the confidence score of the
correct label $y_i$ of a training example $\x_i$ to be larger than the
confidence of other labels.  We define $\Ycal$ as a set of $K$ labels:
$\Ycal=\{1,2,\dots,K\}$. The discriminant function  $ F: \Xcal \times
\Ycal \mapsto \Real $ we need to learn is: $F( \x, y; \w  )  =
\w_y^\T  \bh_y(\x) = {\textstyle \sum}_j w_{(y,j)} \h_{(y,j)}(\x)$.
The class label prediction $y^\star$ for an unknown example $\x$ is to
maximize $F( \x, y; \w  )$ over $y$, which means finding a class label
with the largest confidence:
            $ y^\star = \argmax_y F ( \x, y; \w ) =
            \argmax_y
            \w_y^\T \bh( \x).
            $
            \mcgg is an extension of \mb\cite{Shen2011Direct} for multi-class
classification.
The only difference is that, in \mbb, different classes
share the same set of weak learners $\bh$. In contrast,
each class
associates a separate set of weak learners. We show that
\mcgg learns a more compact model than \mbb.

\begin{algorithm}[t]
\caption{\small CG: Column generation for \mcg}
\footnotesize{
1: {\bf Input:} training examples $ (\x_1; y_1), (\x_2; y_2) ,\cdots
    $; regularization parameter $C$; termination threshold and
    the maximum iteration number.

    2: {\bf Initialize:}  Working
    weak learner set $\Hcal_c = \emptyset ( c = 1\cdots K)$;
    initialize $ \forall (i,y \neq y_i): \lambda_{(i,y)}= 1 $ ($i=1,\dots,m, y=1,\dots,K$).

    3: {\bf Repeat}

    4:$\quad-$ Solve \eqref{eq:colgen} to find $K$ weak learners:
    $\h_c^\star ( \cdot), c = 1\cdots K $;
        and add them to the working weak learner set $\Hcal_c$.

    5:$\quad-$
    Solve the primal problem \eqref{eq:mcgcon} on the current working
    weak learner sets: $\h_c \in \Hcal_c, c = 1,\dots, K $.

    to obtain $ \w $ (we use coordinate
    descent of Algorithm \ref{ALG:alg2}).

    6:$\quad-$
    Update dual variables $ \blambda $ in \eqref{lambda-update}
    using the primal solution $\w$ and the KKT conditions \eqref{lambda-update}.

    7: {\bf Until} the relative change of the primal objective function value
        is smaller than the prescribed tolerance;
        or the maximum iteration is reached.

    8: {\bf Output:}
    $ K $ discriminant function $ F( \x, y; \w  )  =  \w_y^\T
    \bh_y(\x) $, $y = 1 \cdots K$.
}
\label{ALG:alg1}
\end{algorithm}

\textbf{Column generation for \mcg}
    To implement boosting, we need to derive the dual problem of
    \eqref{eq:mcgcon}.
    Similar to \cite{Shen2011Direct},
    the dual problem of \eqref{eq:mcgcon} can be written as
\eqref{eq:mcg-exp-dual}, in which $c$ is the index of class labels. $
\lambda_{ (i,y ) }$ is the dual variable
    associated with one constraint in \eqref{eq:mcgcon}:
    \begin{subequations}
      \label{eq:mcg-exp-dual}
        \begin{align}
        \max_{\blambda}   \;\; &
        \sum_{i} \sum_{y \neq y_i} \lambda_{ (i, y) } \big[ 1 -  \log \tfrac{p}{C} -\log \lambda_{ (  i, y) }  \big]
         \\
        \st \;\; & \forall c=1, \dots, K: \notag\\
& \sum_{i(y_i=c)} \sum_{y \neq y_i } \lambda_{ (i, y)} \bh_{y_i}( \xb_i) -
\sum_{i}  \sum_{y \neq y_i, y=c} \lambda_{ (  i, y)} \bh_y (\xb_i )
  \leq \ones, \label{eq:dualcon}\\
	& \forall i=1, \dots, m: 0 \leq  \sum_{ y \ne y_i } \lambda_{ (
    i, y) } \leq \tfrac{C}{p}.
    \end{align}
    \end{subequations}
Following the idea of column generation \cite{Shen2011Totally},
we divide the original
problem \eqref{eq:mcgcon} into a master problem and a sub-problem,
and solve them alternatively. The master problem is a restricted
problem of \eqref{eq:mcgcon} which only considers the generated weak
learners. The sub-problem is to generate $K$ weak learners
(corresponding $K$ classes) by finding the most violated constraint of
each class in the dual form \eqref{eq:mcg-exp-dual}, and add them to the
master problem at each iteration. The sub-problem for finding most
violated constraints can be written as:
       \begin{align}
            \label{eq:colgen}
    \forall i = &  1 \cdots K:
    \notag \\
    &
    \h_c^\star ( \cdot ) =
    \argmax_{ \h_c(\cdot)}
    \sum_{i(y_i=c)} \sum_{y \neq y_i } \lambda_{ (i, y)}
    \bh_{y_i}( \xb_i) - \sum_{i}  \sum_{y \neq y_i, y=c } \lambda_{ (  i, y)}
    \bh_y (\xb_i ).
	\end{align}
The column generation procedure for \mcg is described in Algorithm
\ref{ALG:alg1}.
Essentially,
we repeat the following two steps until convergence:
1) We solve the
master problem \eqref{eq:mcgcon} with $\h_c \in \Hcal_c, c=1,\dots,K
$, to obtain the primal solution $\w$. $\Hcal_c$ is the working set of generated
weak learners associated with the $c$-th class.
We obtain the dual solution $\blambda^\star$ from the primal solution
$\w^\star$ using the KKT conditions:
\begin{align}
\label{lambda-update}
\lambda_{(i,y)} ^\star =
\frac{C}{p} \exp \big[  \w_y^{\star\T} \bh_y (\xb_i )
-\w_{y_i}^{\star\T}  \bh_{y_i}( \xb_i) \big].
\end{align}
2)
With the dual solution $\lambda_{(i,y)} ^\star$, we solve the
sub-problem \eqref{eq:colgen} to generate $K$ weak learners:
$\h_c^\star, c=1,2,\dots,K $, and add to the working weak learner set
$\Hcal_c$.
	In \mcgg, $K$ weak learners are generated for $K$ classes respectively in each iteration,
    while in \mbb, only one weak learner is generated at each column
    generation and
    shared by all classes. As shown in \cite{Shen2011Direct}
    for \mb, the
    sub-problem for finding the most violated constraint in the dual
    form is:
  \begin{equation}
            \label{eq:colgen_mb}
    [\h^\star ( \cdot ), \; c^\star]= \argmax_{ \h(\cdot), \; c}
    \sum_{i(y_i=c)} \sum_{y \neq y_i }
    \lambda_{ (i, y)} \bh( \x_i) - \sum_{i}  \sum_{y \neq y_i, y=c } \lambda_{ (
    i, y)} \bh (\x_i ).
   \end{equation}
        At each column generation of \mbb, \eqref{eq:colgen_mb} is
        solved to generated one weak learner. Note that solving
        \eqref{eq:colgen_mb} is to search over {\em all $K$ classes}
        to find the
        best weak learner $\h^\star$. Thus the computational cost is
        the same as \mcgg. This is the reason why \mcg does not introduce
        additional training cost compared to \mb.
        In general, the solution
        $[\w_1; \cdots; \w_K ]$ of \mb is highly sparse
        \cite{Shen2011Direct}.
         This can be observed in
        our empirical study. The weak learner generated by solving
        \eqref{eq:colgen_mb} actually is targeted for one class, thus
        using this weak learner across all classes in \mb leads to a
        very sparse solution.
    The sparsity of $[\w_1,\cdots,\w_K]$ indicates that one weak
    learner is usually only useful for the prediction of a very few number of
    classes (typically only one), but useless for most other classes.
    In this sense, forcing different classes to use the same set of
    weak learners may not be necessary and usually it leads to slow
    convergence. In contrast, using separate weak learner sets for
    each class, \mcg tends to have a dense solution of $\w$.
    With $ K $ weak learners generated at each iteration, \mcg
    converges much faster.

{\bf Fast coordinate descent}
To further speed up the training, we propose a fast coordinate descent
    method (\scd) for solving the primal \mcg problem
    at each column generation iteration.
The details of \scd is presented in Algorithm \ref{ALG:alg2}.
    The high-level idea is simple.
    \scd works iteratively, and
    at each iteration (working set iteration), we compute the violated
    value of the KKT conditions for each variable in $\w$, and
    construct a working set of violated variables (denoted as $\Scal$), then
    pick variables from the $\Scal$ for update (one variable at a
    time). We also use the violated values for defining stop criteria.
    Our \scd is
    a mix of sequential and stochastic coordinate descent. For the
    first working set iteration, variables are sequentially picked for
    update (cyclic CD); in
    later working set iterations, variables are randomly picked (stochastic CD).
    In the sequel, we present the details of \scd.
        First, we describe how to update one variable of $\w$ by
        solving a single-variable sub-problem. For notation simplicity,
        we define:
$
        \dh_i(y)= \bh_{y_i}(\x_i) \otimes \Gamma (y_i) - \bh_y(\x_i)
        \otimes \Gamma (y).
$
$\Gamma(y)$ is the orthogonal label coding vector:
$    \Gamma (y) = [ \delta(y,1), \delta(y,2), \cdots, \delta(y, K)
]^\T $ $\in \{0, 1\}^K$. Here  $\delta(y,k)$ is the indicator function that
returns 1 if $y=k$, otherwise $0$. $ \otimes $ denotes the tensor
product. \mcg in \eqref{eq:mcgcon} can be equivalently written as:
\begin{align}
	\label{eq:mcg-exp2}
          \min_{\w \geq 0}   \;\; &
  \|\w \|_1 +  {\tfrac{C}{p}} \, \sum_{i} \sum_{y\neq y_i}  \exp
  \big[- \w^\T \dh_i(y)\big].
\end{align}
    We assume that binary weak learners are used here: $\h(\x) \in \{
    +1, -1\} $.  $\dhh_{i,j}(y)$ denotes the $j$-th dimension of
    $\dh_i(y)$, and $\dhat_{i,j}(y)$ denotes the rest dimensions of
    $\dh_i(y)$ excluding the $j$-th. The output of $\dhh_{i,j}(y)$
    only takes three possible values: $\dhh_{i,j}(y) \in \{-1, 0, +1
    \}$. For the $j$-th dimension, we define: $\Dcal_v^j=\{\; (i, y)\;
    | \;\dhh_i^j(y) =  v,
i\in\{1,\dots,m\}, y\in\Ycal/y_i \; \}, v \in \{-1, 0, +1 \}$; so
$\Dcal_v^j$ is a set of constraint indices $(i,y)$ that the output of
$\dhh_{i,j}(y)$ is $v$. $w_j$ denotes the $j$-th variable of $\w$;
$\wh_j$ denotes the rest variables of $\w$ excluding the $j$-th. Let
$g(\w)$ be the objective function of the optimization
\eqref{eq:mcg-exp2}. $g(\w)$ can be de-composited as:
\begin{align}
  g(\w) & = \|\w \|_1 +  {\tfrac{C}{p}} \sum_{i} \sum_{y\neq y_i} \exp \big[-  \w^\T \dh_i(y)\big] \notag  \\
  & = \|\wh_j \|_1 + \|w_j \|_1 + {\tfrac{C}{p}} \sum_{i, y\neq y_i}
  \exp \big[- \wh_j^\T \dhat_{i,j}(y) - w_j^\T \dhh_{i,j}(y) \big]
  \notag \\
  & = \|\wh_j \|_1 + \|w_j \|_1 +  {\tfrac{C}{p}} \bigg\{ \exp(w^\T_j) \sum_{(i, y)\in \Dcal_{-1}^j} \exp\big[- \wh_j^\T \dhat_{i,j}(y)\big] + \notag \\
  & \quad \exp(-w^\T_j) \sum_{(i, y)\in \Dcal_{+1}^j} \exp \big[- \wh_j^\T \dhat_{i,j}(y)\big] +
  \sum_{(i, y)\in \Dcal_{0}^j} \exp \big[- \wh_j^\T \dhat_{i,j}(y)\big]
    \bigg\} \notag \\
  & = \|\wh_j \|_1 + \|w_j \|_1 +  {\tfrac{C}{p}} \big[\exp(w^\T_j)
  \bV_- + \exp(-w^\T_j) \bV_+ + \bV_0 \big].
  \label{eq:scd_g}
\end{align}
Here we have defined:
\begin{subequations}
\begin{align}
\bV_-=\sum_{(i, y)\in \Dcal_{-1}^j} \exp \big[- \wh_j^\T \dhat_{i,j}(y)\big],  & \quad
\bV_0=\sum_{(i, y)\in \Dcal_{0}^j} \exp \big[- \wh_j^\T \dhat_{i,j}(y)\big], \label{eq:V}\\
\bV_+=\sum_{(i, y)\in \Dcal_{+1}^j} \exp \big[- \wh_j^\T \dhat_{i,j}(y)\big]. \label{eq:V2}
\end{align}
\end{subequations}
    In the variable update step, one variable $w_j$ is picked at a
    time for updating and other variables $\wh_j$ are fixed; thus we
    need to minimize $g$ in \eqref{eq:scd_g} w.r.t $w_j$, which is a
    single-variable minimization. It can be written as:
\begin{align}
	\label{eq:mcg-exp3}
          \min_{w_j \geq 0}   \;\; &
 \|w_j \|_1 +  {\tfrac{C}{p}} \big[ \bV_- \exp(w^\T_j) + \bV_+
 \exp(-w^\T_j) \big].
\end{align}
The derivative of the objective function in \eqref{eq:mcg-exp3} with $w_j > 0$ is:
 \begin{align}
 \label{eq:scd-sub}
\frac{\partial g}{\partial w_j} = 0  \implies & 1 + \tfrac{C}{p} \big[
\bV_- \exp(w^\T_j) - \bV_+ \exp(-w^\T_j) \big] = 0.
 \end{align}
 By solving \eqref{eq:scd-sub} and the bounded constraint $w_j\geq0$,
 we obtain the analytical solution of the optimization in
 \eqref{eq:mcg-exp3} (since $\bV_->0$):
\begin{align}
	\label{eq:scd-w}
	w_j^\star= \max \bigg\{ 0, \;\; \log \bigg(\sqrt{ \bV_+ \bV_- +
    \tfrac{p^2}{4C^2}} - \tfrac{p}{2C} \bigg) -\log{\bV_-} \bigg\}.
\end{align}
    When $C$ is large,
    \eqref{eq:scd-w} can be approximately simplified as:
\begin{align}
	\label{eq:scd-w2}
	w_j^\star=  \max \bigg\{ 0, \;\; \frac{1}{2} \log \frac{\bV_+}{
    \bV_-} \bigg\}.
\end{align}
With the analytical solution in \eqref{eq:scd-w}, the update of each dimension of $\w$ can be performed extremely efficiently.
    The main requirement for obtaining the closed-form solution is
    that the use of discrete weak learners.

We use the KKT conditions to construct a set of violated variables and derive meaningful stop criteria. For the optimization of \mcgg
\eqref{eq:mcg-exp2}, KKT conditions are necessary conditions and also
sufficient for optimality. The Lagrangian of \eqref{eq:mcg-exp2} is:
  $
  L =
  \|\w \|_1 +  {\tfrac{C}{p}} \, \sum_{i} \sum_{y\neq y_i}  \exp
  \big[- \w^\T \dh_i(y)\big]  - \balpha^\T \w.
  $
According to the KKT conditions, $\w^\star$ is the optimal for
\eqref{eq:mcg-exp3} if and only if $\w^\star$ satisfies $\w^\star \geq
0, \balpha^\star \geq 0, \forall j: \alpha^\star_jw^\star_j=0$ and $
\forall j: \nabla_j L(\w^\star)=0$. For $w_j > 0$,
 \begin{align}
\frac{\partial L}{\partial w_j} = 0  \implies &
 1 -  {\tfrac{C}{p}} \, \sum_{i} \sum_{y\neq y_i}  \exp \big[- \w
 \dh_{i}(y)\big] \dhh_{i,j}(y) - \alpha_j = 0.  \notag
 \end{align}
Considering the complementary slackness: $\alpha^\star_jw^\star_j=0$,
if $w^\star_j>0$, we have $\alpha^\star_j=0$; if $w^\star_j=0$, we
have $\alpha^\star_j \geq 0$. The optimality conditions can be written
as:
\begin{equation}
	\label{eq:kkt}
  \forall j: \left\{
  \begin{array}{l l}
1 -  {\tfrac{C}{p}} \, \sum_{i} \sum_{y\neq y_i}  \exp \big[- \w^\star
\dh_{i}(y)\big] \dhh_{i,j}(y) = 0,
    & \quad \text{if $w_j^\star > 0$};\\
1 -  {\tfrac{C}{p}} \, \sum_{i} \sum_{y\neq y_i}  \exp \big[- \w^\star
\dh_{i}(y)\big] \dhh_{i,j}(y) \geq 0,
    & \quad \text{if $w_j^\star = 0$}.\\
  \end{array} \right.
\end{equation}
For notation simplicity, we define a column vector $\bmu$ as in \eqref{eq:mu}.
With the optimality conditions \eqref{eq:kkt}, we define $\btheta_j$
in \eqref{eq:vio} as the violated value of the $j$-th variable of the
solution $\w^\star$:
\begin{align}
\label{eq:mu}
\bmu_{(i,y)}= \exp \big[- \w^\T \dh_i(y)\big]
\end{align}
\begin{equation}
\label{eq:vio}
  \btheta_j= \left\{
  \begin{array}{l l}
  |1 -  {\tfrac{C}{p}} \, \sum_{i} \sum_{y\neq y_i}  \bmu_{(i,y)} \,
  \dhh_{i,j}(y) |
    & \quad \text{if $w_j^\star > 0$}\\
 \max \{ 0, \; {\tfrac{C}{p}} \, \sum_{i} \sum_{y\neq y_i} \bmu_{(i,y)} \, \dhh_{i,j}(y) - 1 \}
    & \quad \text{if $w_j^\star = 0$}.\\
  \end{array} \right.
\end{equation}
At each working set iteration of \scd, we compute the violated values $\btheta$,
and construct a working set $\Scal$ of violated variables; then we
randomly (except the first iteration) pick one variable from $\Scal$
for update. We repeat picking for $|\Scal|$ times; $|\Scal|$ is the
element number of $\Scal$. $\Scal$ is defined as
\begin{equation}
 \label{eq:workset}
\Scal = \{j\,|\, \btheta_j > \epsilon \;\}
\end{equation}
where $\epsilon$ is a
tolerance parameter.
Analogue to \cite{Fan2008} and \cite{Yuan2010JMLR}, with the
definition of the variable violated values $\btheta$ in
\eqref{eq:vio}, we can define the stop criteria as:
\begin{align}
\label{eq:scd-stop}
\max_j \btheta_j \leq \epsilon,
\end{align}
where $ \epsilon$ can be the same tolerance parameter as in the
working set $ \Scal$ definition \eqref{eq:workset}.
The stop condition \eqref{eq:scd-stop} shows if the largest violated
value is smaller than some threshold, \scd terminates. We  can see
that using KKT conditions is actually using the gradient information.
    An inexact solution for $\w$ is acceptable for each column generation
iteration, thus we place a maximum iteration number ($\Z$ in Algorithm 2) for  \scd to prevent
unnecessary computation.
We need to compute $\bmu$ before obtaining $\btheta$, but computing
$\bmu$ in \eqref{eq:mu} is expensive. Fortunately, we are able to
efficiently update $\bmu$ after the update of one variable $w_j$ to
avoid re-computing of \eqref{eq:mu}. $\bmu$ in \eqref{eq:mu} can be
equally written as:
\begin{align}
\label{eq:mu2}
\bmu_{(i,y)} = \exp \big[- \wh_j^\T \dhat_{i,j}(y) - w_j \dhh_{i,j}(y)
\big].
\end{align}
So the update of $\bmu$ is then:
\begin{align}
\label{eq:mu-update}
\bmu_{(i,y)}=\bmu_{(i,y)}^\text{old} \exp \big[
\dhh_{i,j}(y)(w_j^{\text{old}} - w_j) \big].
\end{align}
With the definition of $\bmu$ in \eqref{eq:mu2}, the values $V_-$ and
$V_+$ for one variable update can be efficiently computed by using
$\bmu$ to avoid the expensive computation in \eqref{eq:V} and
\eqref{eq:V2}; $V_-$ and $V_+$ can be equally defined as:
\begin{align}
\label{eq:V3}
\bV_-=\sum_{(i, y)\in \Dcal_{-1}^j} \bmu_{(i,y)} \exp(-w_j),  & \quad
\bV_+=\sum_{(i, y)\in \Dcal_{+1}^j} \bmu_{(i,y)} \exp(w_j).
\end{align}

Some discussion on \scd ({Algorithm \ref{ALG:alg2}}) is as follows:
1) Stage-wise optimization is a special case of \scd.
Compared to totally corrective optimization which considers all
variables of $\w$ for update, stage-wise only considers those newly
added variables for update. We initialize the working set using the
newly added variables. For the first working set iteration, we sequentially update
the new added variables. If setting the maximum working set iteration to $1$ ($\Z=1$ in Algorithm 2),
\scd  becomes a stage-wise algorithm. Thus \scd is a
generalized algorithm with totally corrective update and stage-wise
update as
special cases. In the stage-wise setting, usually a large $C$
(regularization parameter) is implicitly enforced, thus we can use the
analytical solution in \eqref{eq:scd-w2} for variable update.

2) Randomly picking one variable for update without any guidance leads
to slow local convergence. When the solution gets close to the
optimality, usually only very few variables need update, and most
picks do not ``hit''. In column generation (CG), the initial
value of $\w$ is initialized by the solution of last CG iteration.
This initialization is already fairly close to optimality. Therefore the slow local
convergence for stochastic coordinate decent (CD) is more serious in column generation
based boosting. Here we have used the KKT conditions to iteratively construct a
working set of violated variables, and only the
variables in the working set need update. This strategy leads to
faster CD convergence.

\begin{algorithm}[t]
\caption{\small \scd: Fast coordinate decent for \mcg}
\footnotesize{
1: {\bf Input:} training examples $ (\x_1; y_1), \cdots, (\x_m; y_m)
    $; parameter $C$; tolerance:
$\epsilon$
    ; weak learner set $\Hcal_c, c = 1,\dots, K$; initial value of $\w$; maximum working set iteration: $\Z$.

    2: {\bf Initialize:}  initialize variable working set $\Scal$ by variable indices in $\w$ that correspond to newly added weak learners; initialize $\bmu$ in \eqref{eq:mu}; working set iteration index $\z=0$.

    3: {\bf Repeat} (working set iteration)

    4:$\quad \quad$ $\z=\z+1$; reset the inner loop index: $q=0$;

    5:$\quad \quad$ {\bf While $q<|\Scal|$ } ($|\Scal|$ is the size of $\Scal$)

    6:$\quad \quad \quad \quad$ $q=q+1;$     pick one variable index $j$ from $\Scal$:

    $\quad \quad \quad \quad \quad \quad \quad \quad $
    if $\z=1$ sequentially pick one, else randomly pick one.

    7:$\quad \quad \quad \quad $
    Compute $V_-$ and $V_+$ in \eqref{eq:V3} using $\bmu$.

    8:$\quad \quad \quad\quad$
	update variable $w_j$ in \eqref{eq:scd-w} using $V_-$ and $V_+$.

    9:$\quad \quad \quad \quad $
	update $\bmu$ in \eqref{eq:mu-update} using the updated $w_j$.

	10:$\quad \quad$ {\bf End While}

	11:$\quad \quad$
	Compute the violated values $\btheta$ in \eqref{eq:vio} for all variables.

	12:$\quad \quad$ Re-construct the variable working set $\Scal$ in \eqref{eq:workset} using $\btheta$.

    13: {\bf Until} the stop condition in \eqref{eq:scd-stop} is satisfied or maximum working set iteration reached: $\z>=\Z$.

    14: {\bf Output:}
     $ \w $.
}
\label{ALG:alg2}
\end{algorithm}

\section{Experiments}

We evaluate our method \mcg on some UCI datasets and a variety of
multi-class image
 classification applications, including digit recognition,
scene recognition, and traffic sign recognition.
We compare \mcg against \mb \cite{Shen2011Direct} with
the exponential loss, and another there popular multi-class boosting
algorithms: AdaBoost.ECC \cite{Guruswami1999},
AdaBoost.MH \cite{Schapire99improvedboosting} and
AdaBoost.MO \cite{Schapire99improvedboosting}. We use \scd as the
solver for \mcgg, and LBFGS-B \cite{lbfgs} for \mbb.
We also perform further experiments to evaluate \scd in detail.
For all experiments, the best regularization parameter $C$ for \mcg
and \mb is selected from $10^2$ to $10^5$;
the tolerance parameter in \scd is set to
$0.1$ ($\epsilon =0.1$); We use \mcgone to denote \mcg using the
stage-wise setting of \scd which only uses one iteration ($\Z=1$ in Algorithm \ref{ALG:alg2}). In
\mcgone, we fix $C$ to be a large value: $C=10^8$.

All experiments are run 5 times. We compare the testing error, the
total training time and solver time on all datasets. The results show
that our \mcg and \mcgone converge much faster then other
methods, use less training time then \mbb, and achieve the best
testing error on most datasets.

\begin{figure}[t]
    \centering
    \subfloat{
	    \includegraphics[width=.33\linewidth]{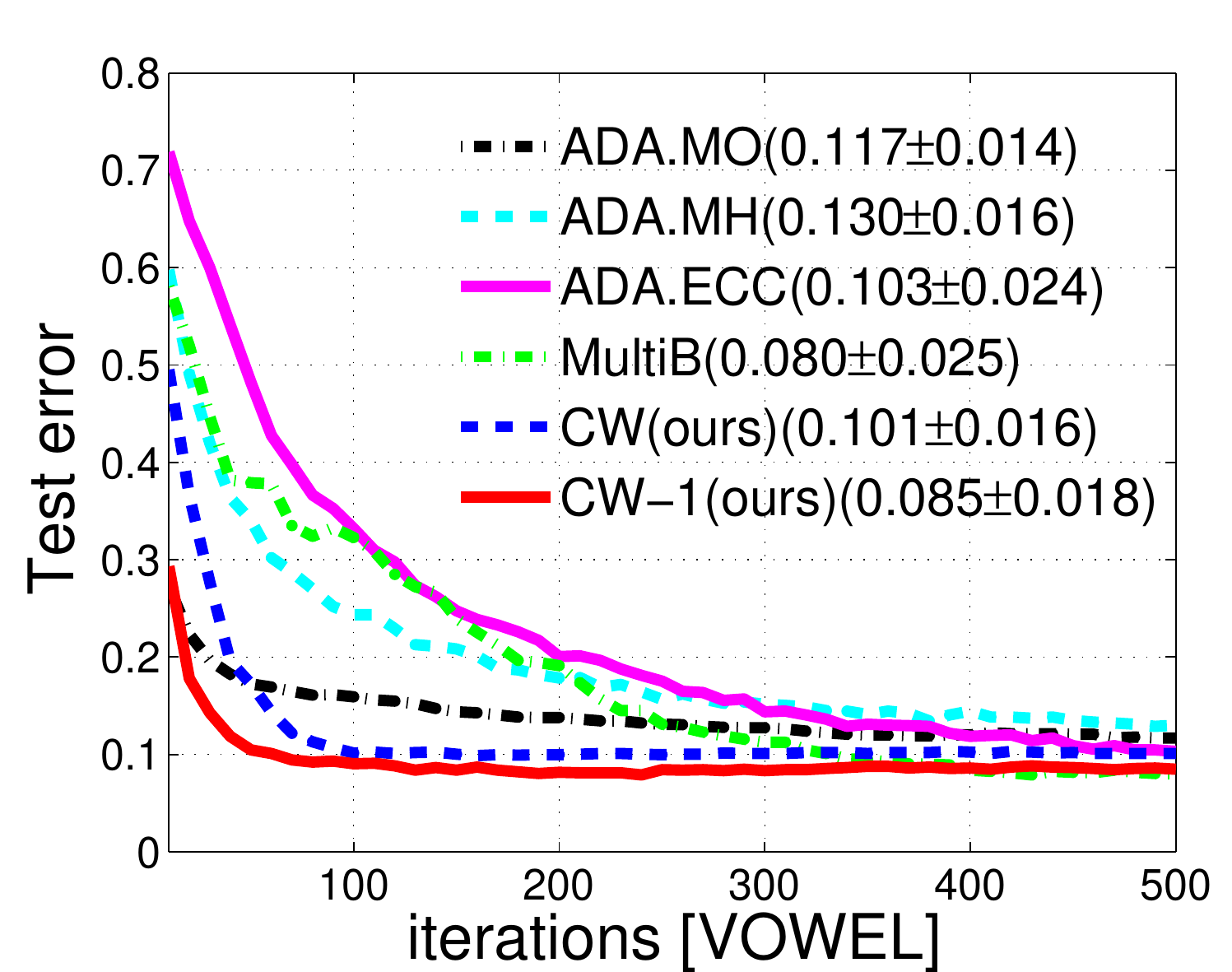}
	    \includegraphics[width=.33\linewidth]{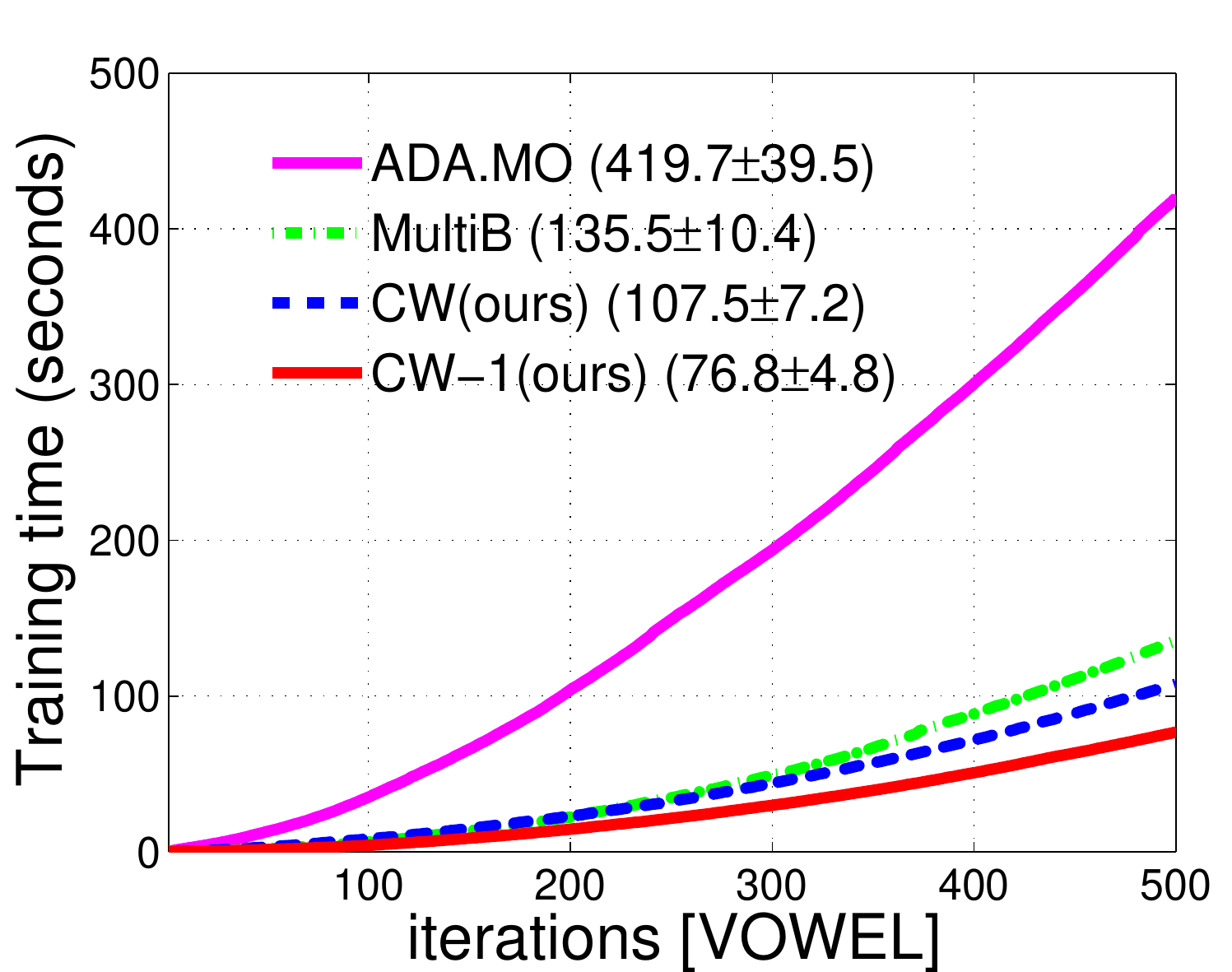}
   	    \includegraphics[width=.33\linewidth]{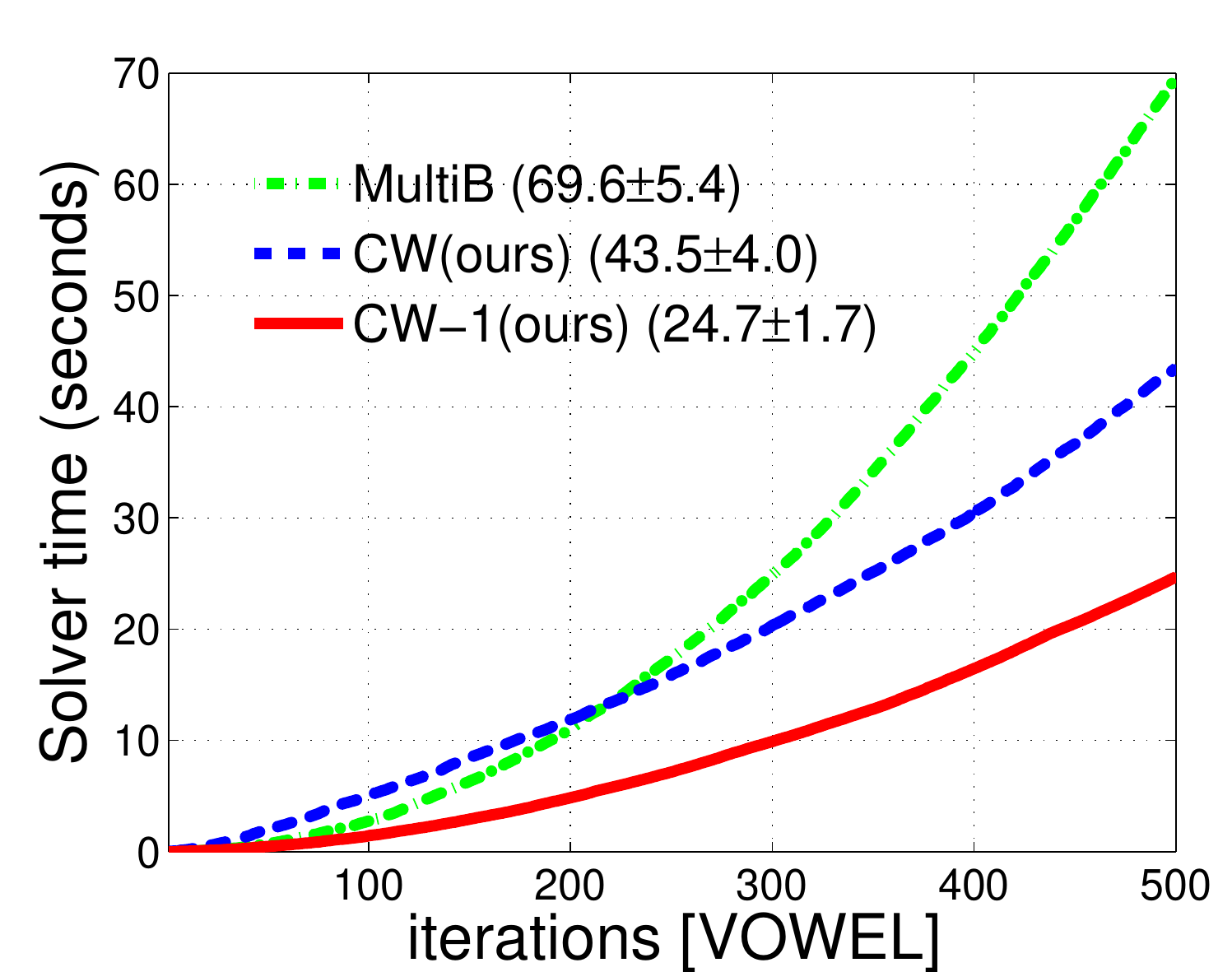}
    }

    \subfloat{
	    \includegraphics[width=.33\linewidth]{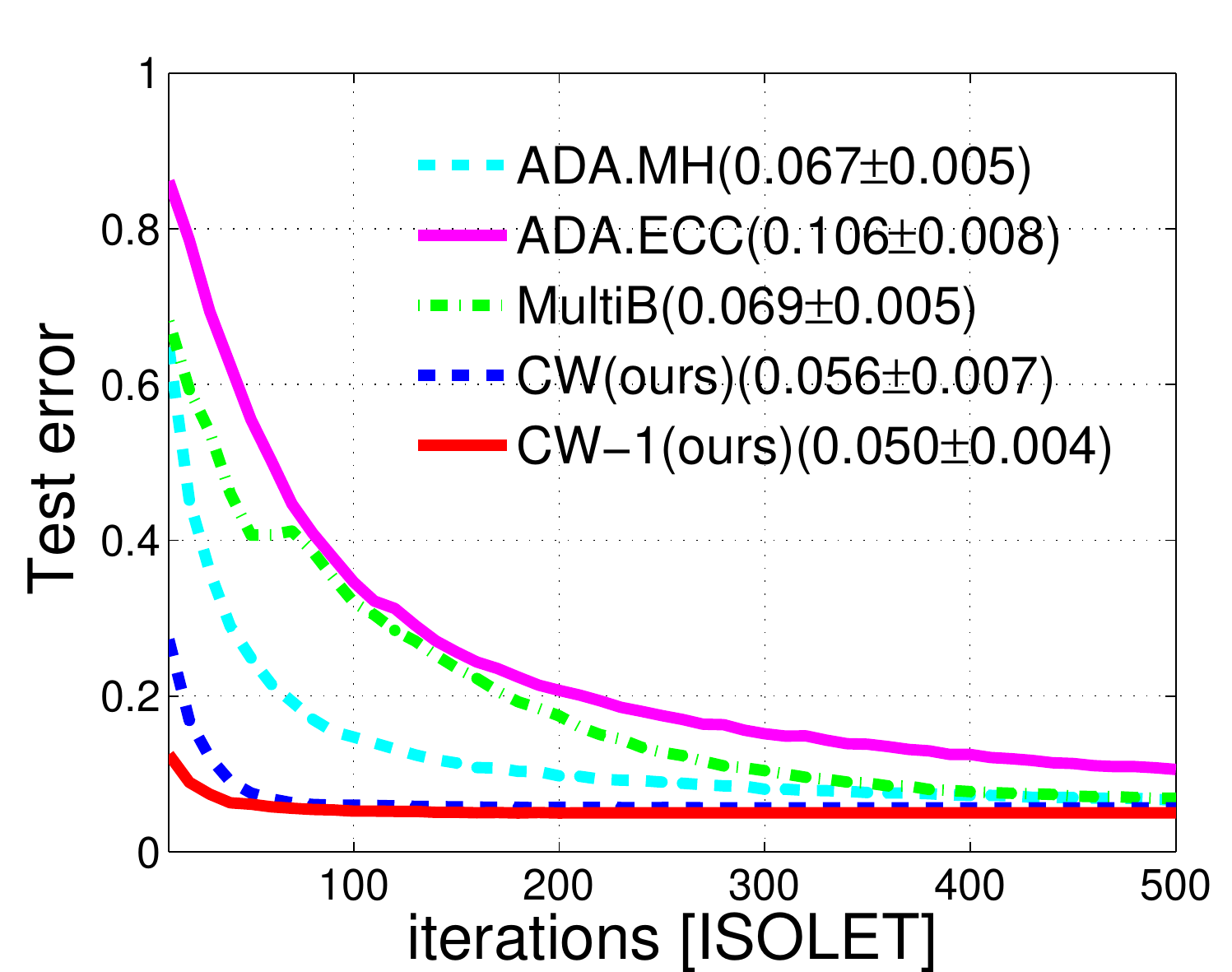}
	    \includegraphics[width=.33\linewidth]{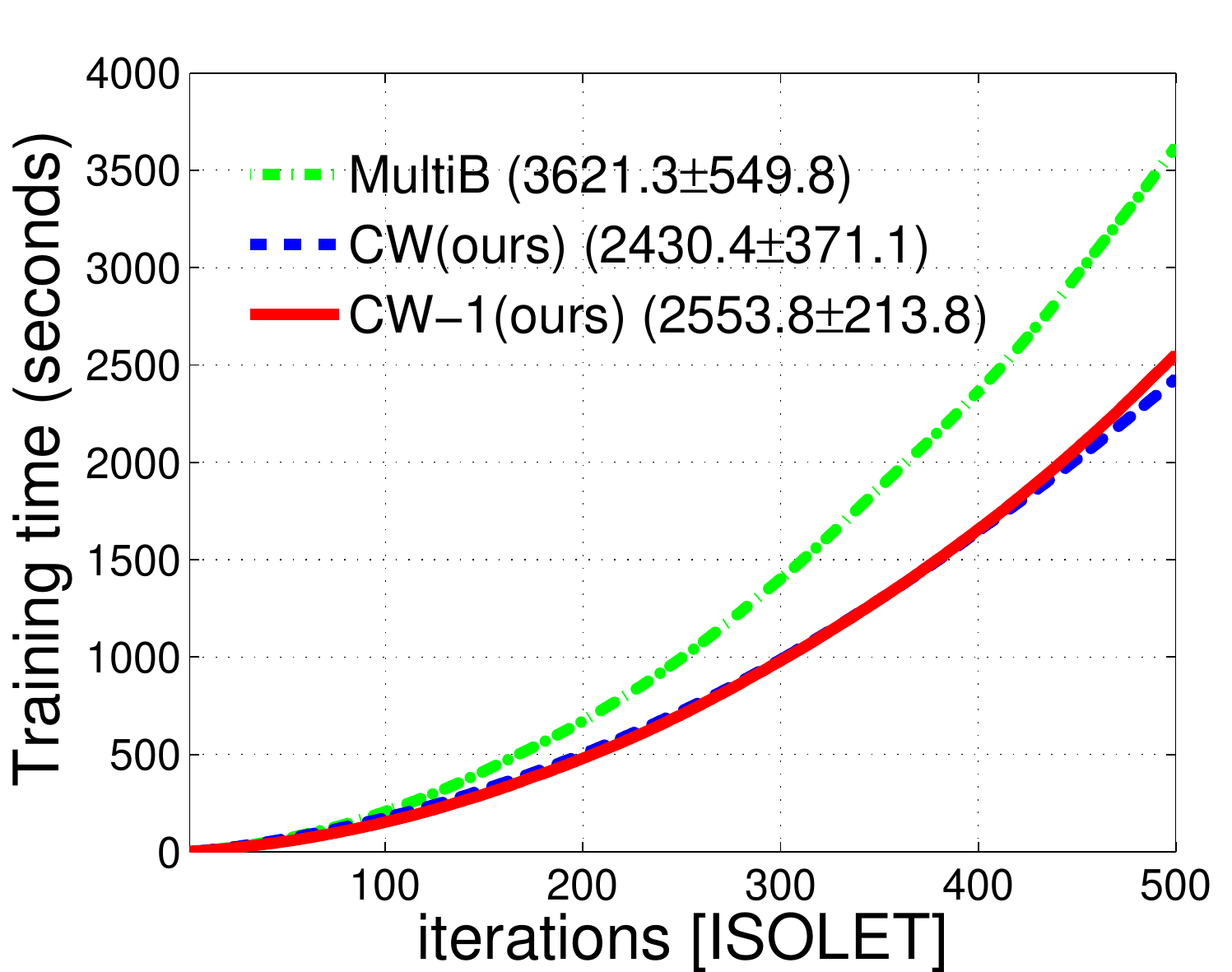}
   	    \includegraphics[width=.33\linewidth]{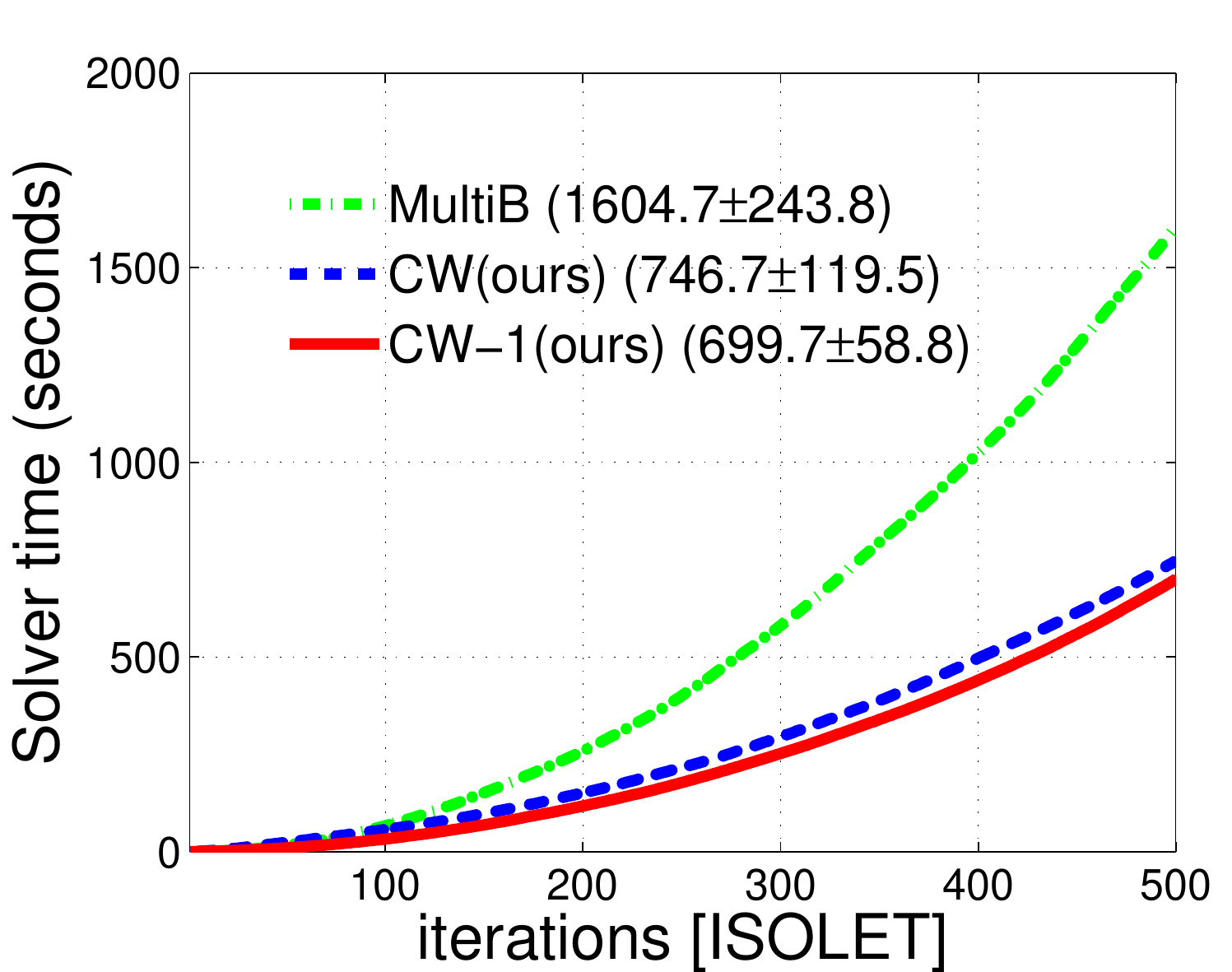}
    }

    \caption{Results of 2 UCI datasets: VOWEL and ISOLET. CW and CW-1
    are our methods. CW-1 uses stage-wise setting. The number after
    the method name is the mean value with standard deviation of the
    last iteration. Our methods converge much faster
    and achieve competitive test accuracy. The total training time and
    the solver time of our methods both are less than MultiBoost of
    \cite{Shen2011Direct}.
    }
    \label{fig:uci}
\end{figure}

\begin{figure}[t]
    \centering
    \subfloat{
    	    \includegraphics[width=.33\linewidth]{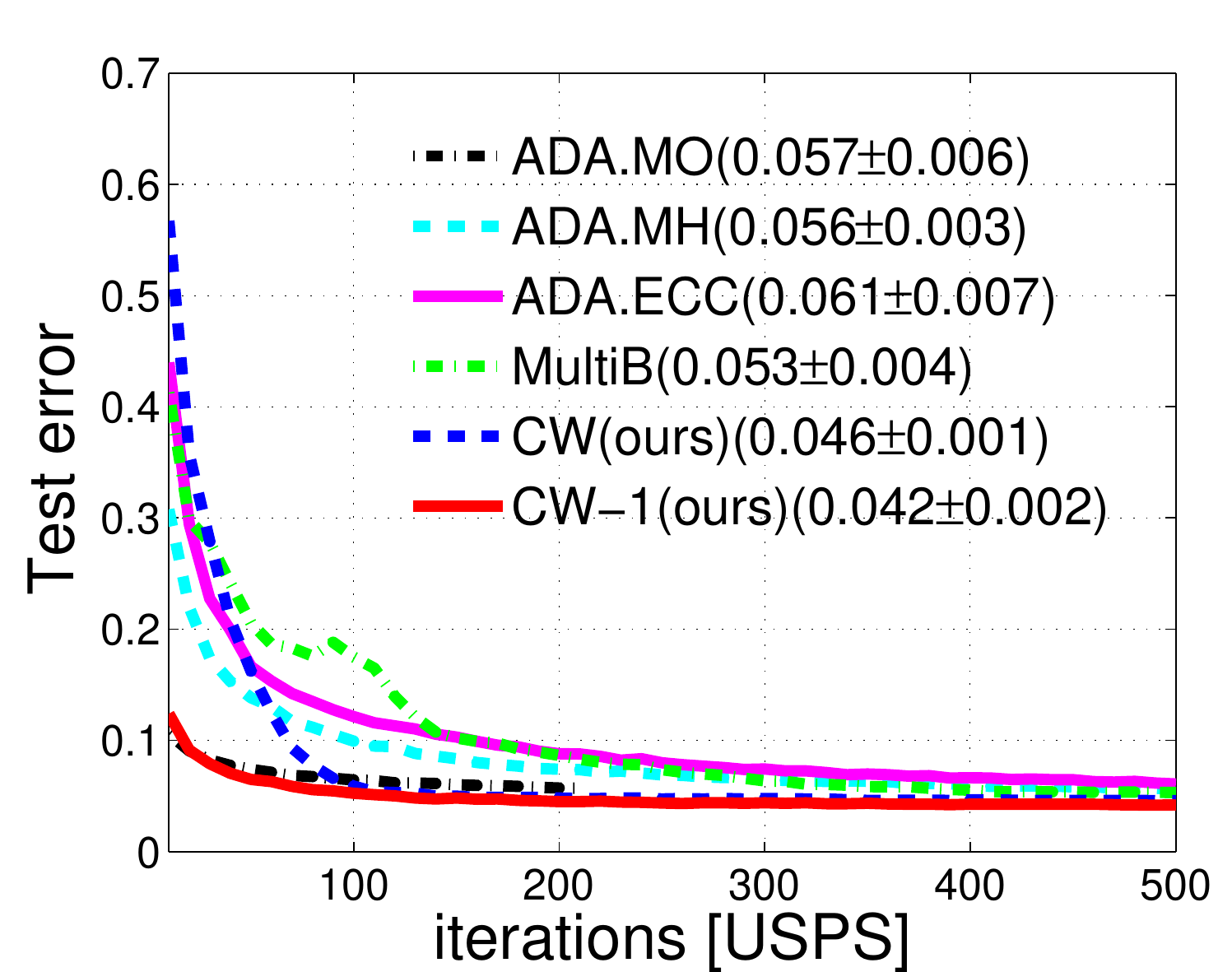}
    	    \includegraphics[width=.33\linewidth]{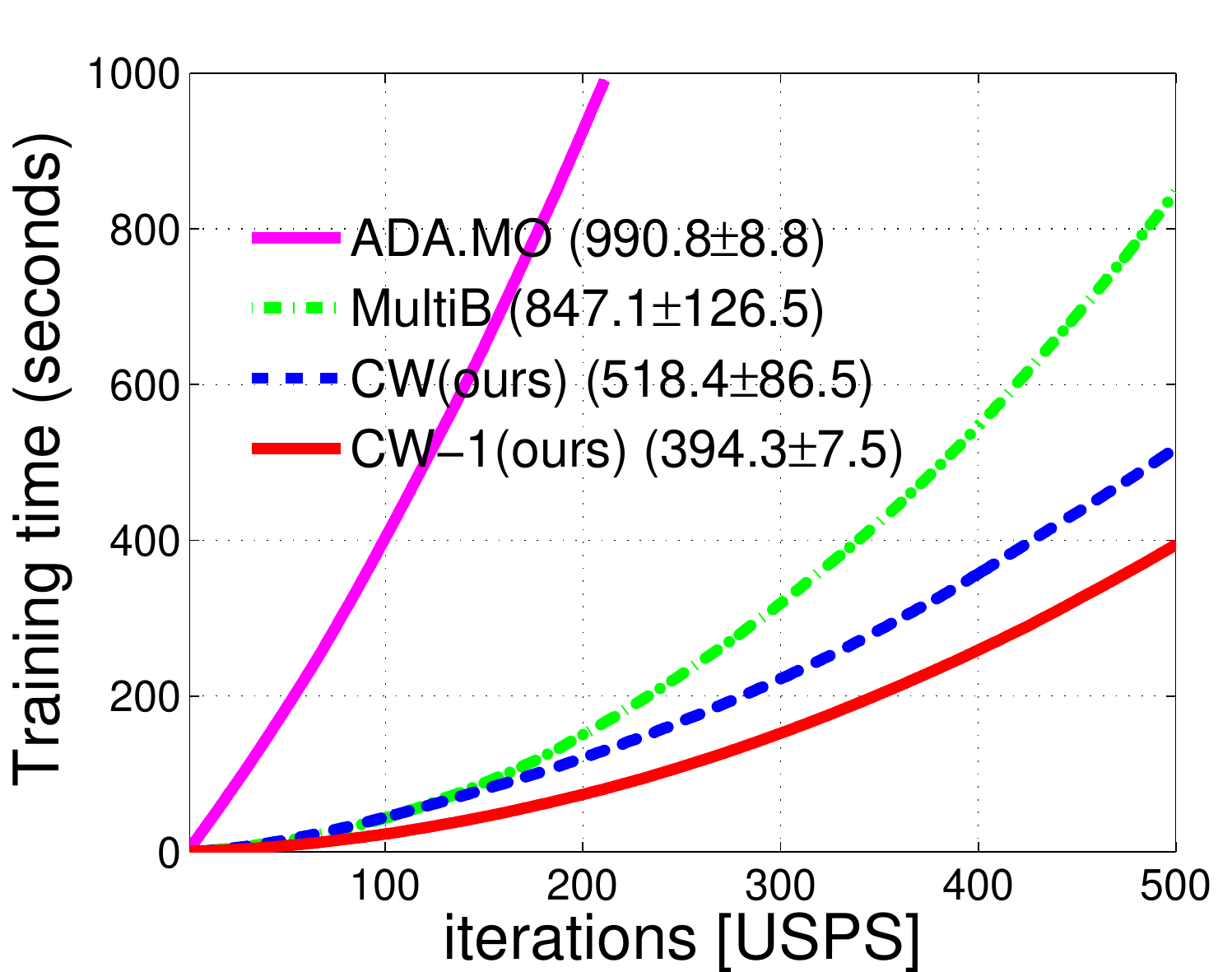}
    	    \includegraphics[width=.33\linewidth]{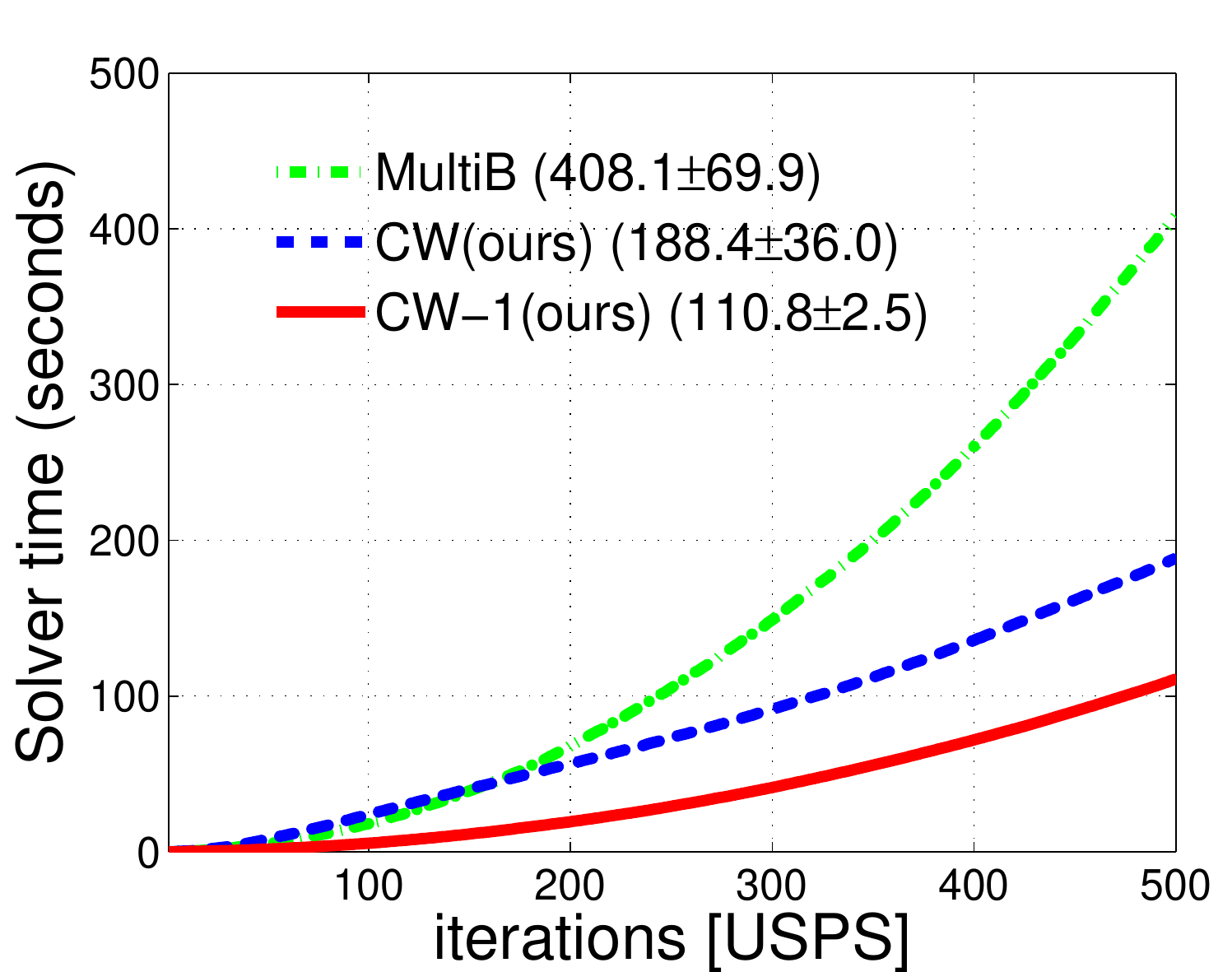}
    }

    \subfloat{
	    \includegraphics[width=.33\linewidth]{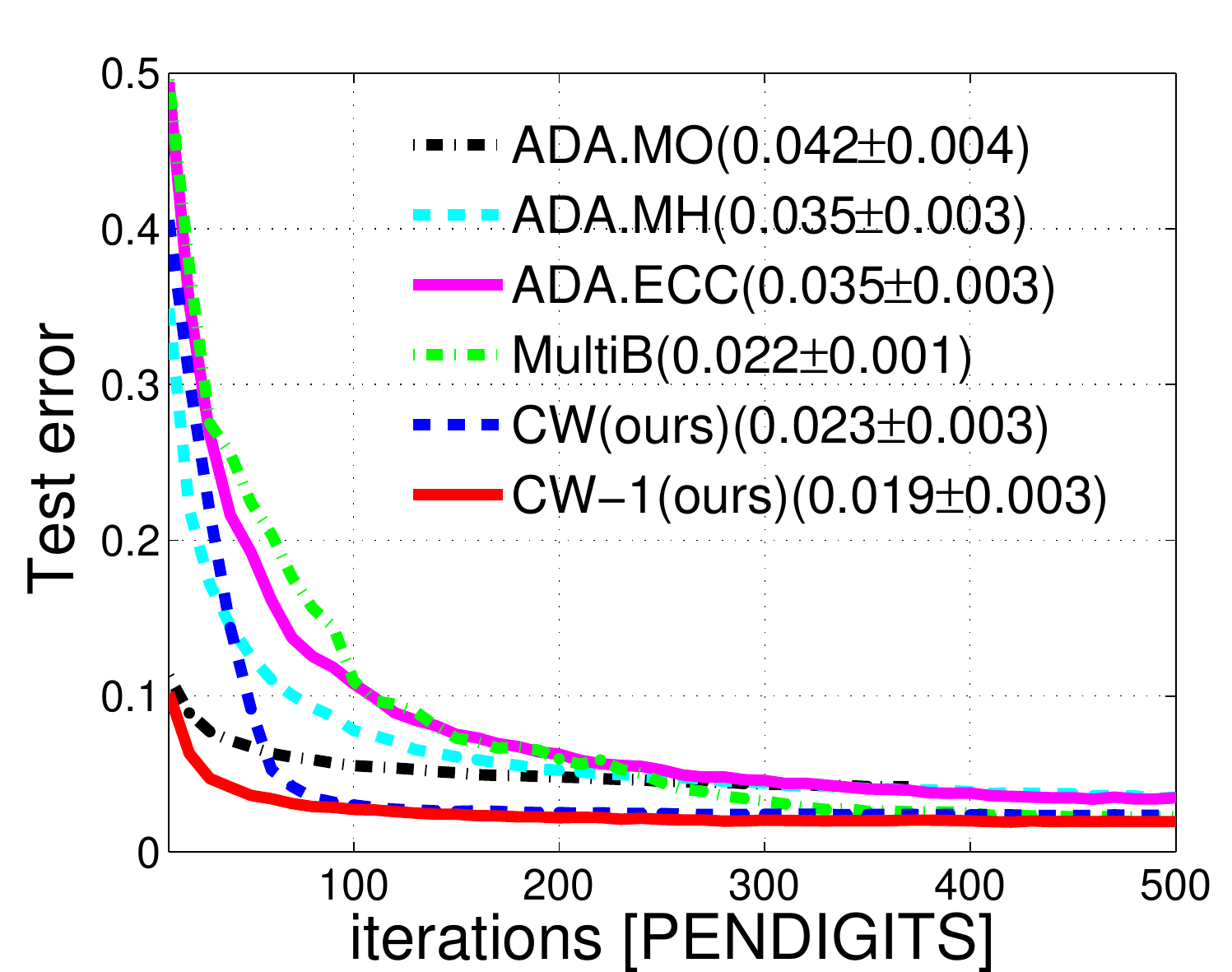}
	    \includegraphics[width=.33\linewidth]{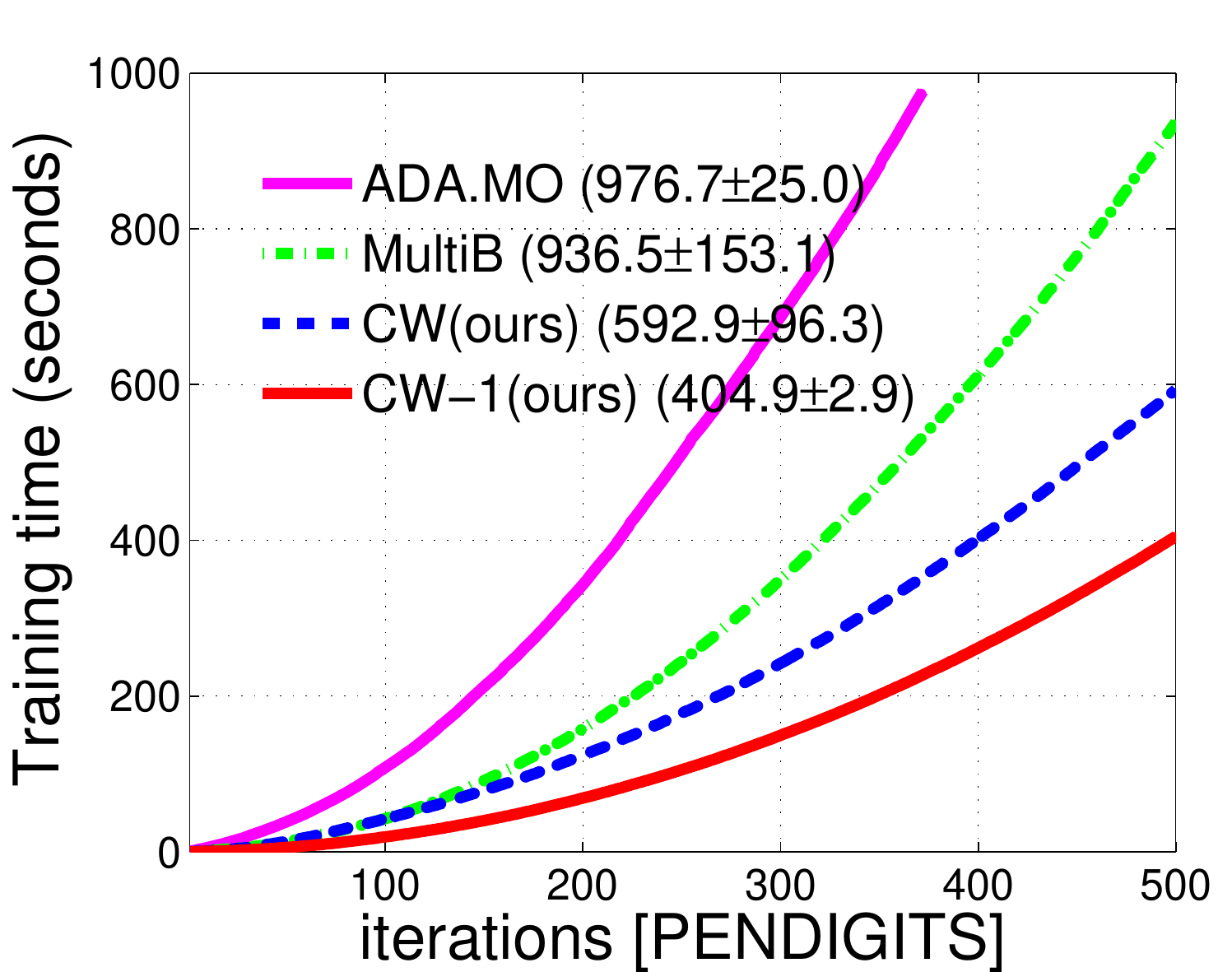}
	    \includegraphics[width=.33\linewidth]{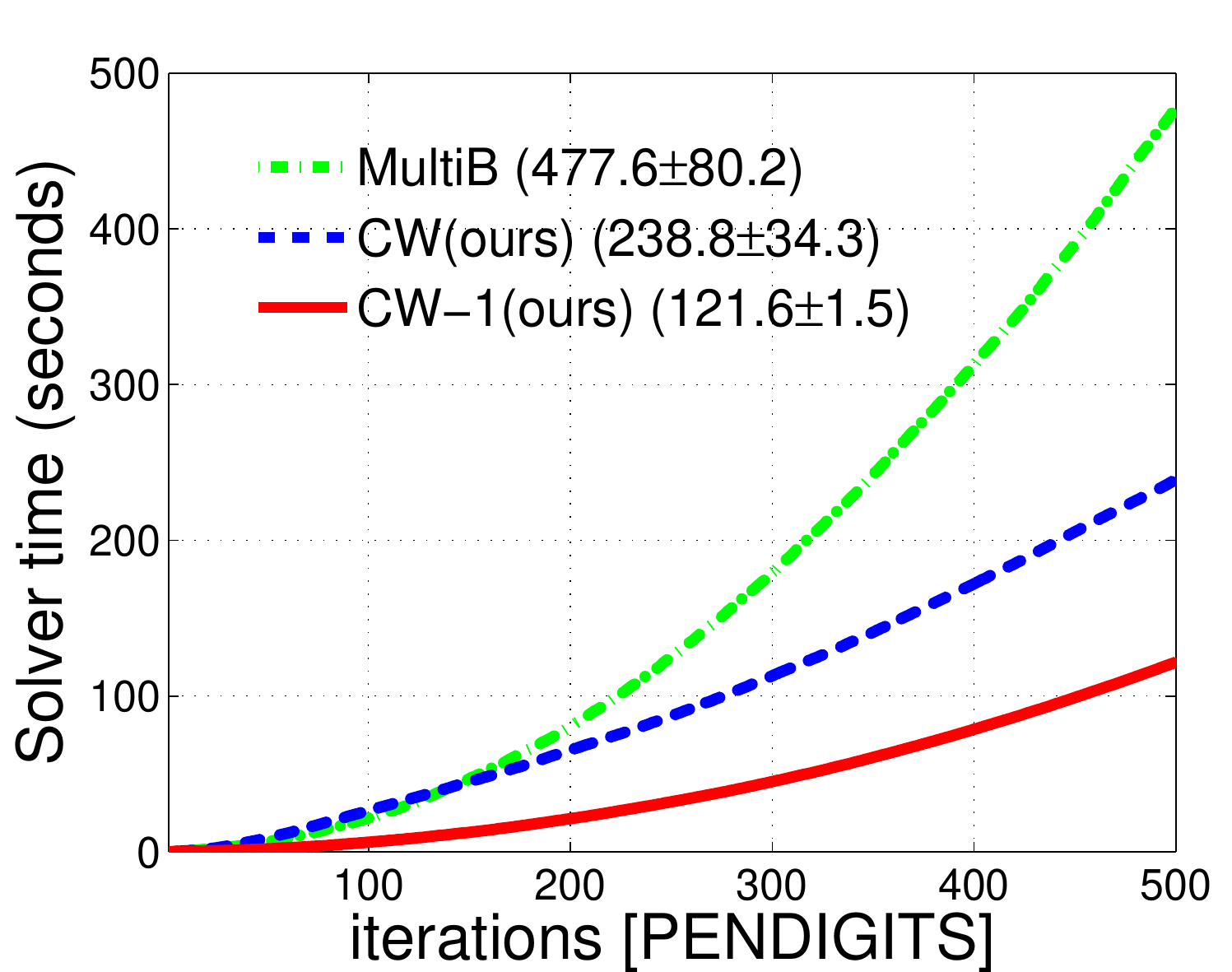}
    }

   \subfloat{
   	    	    \includegraphics[width=.33\linewidth]{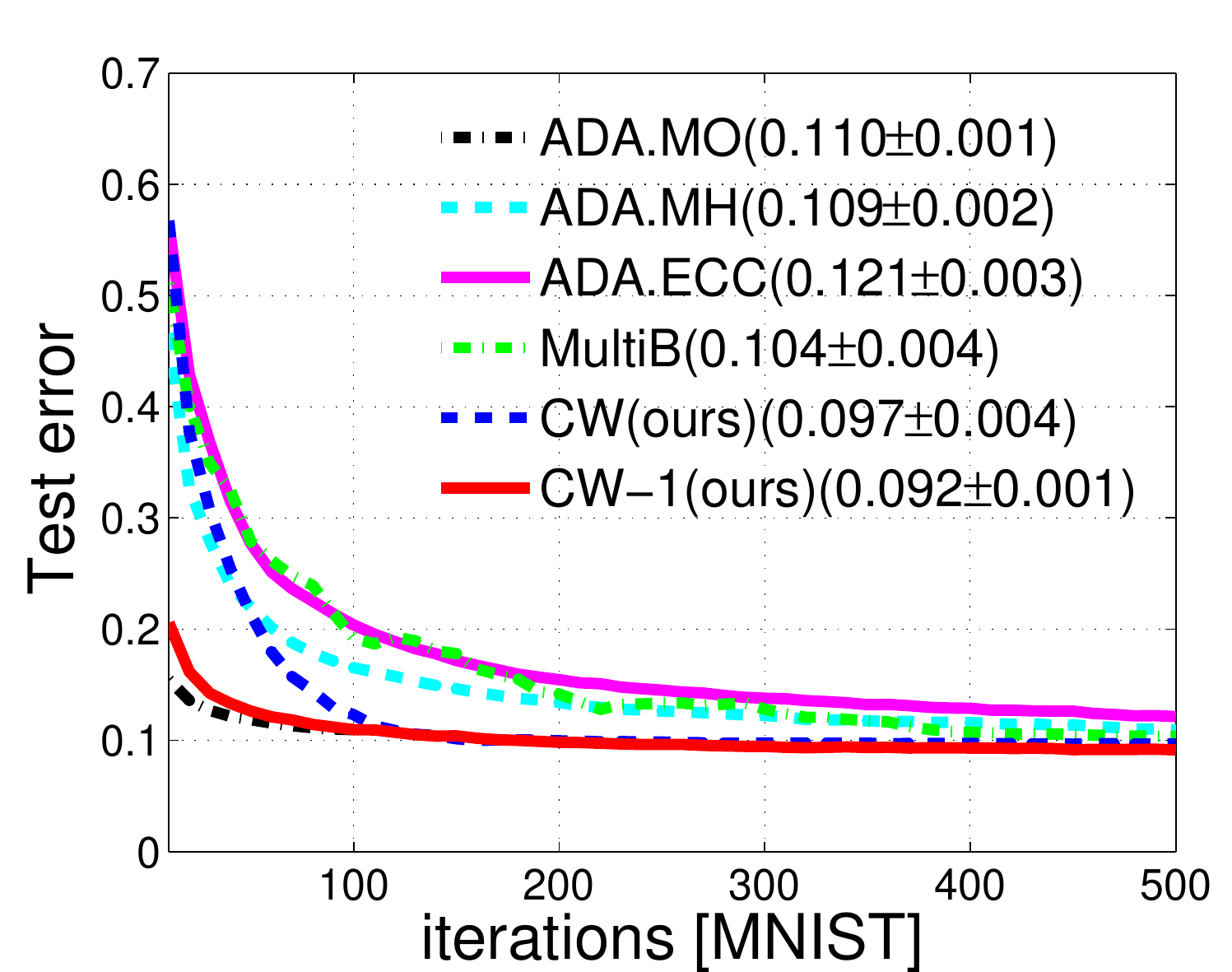}
   	    	    \includegraphics[width=.33\linewidth]{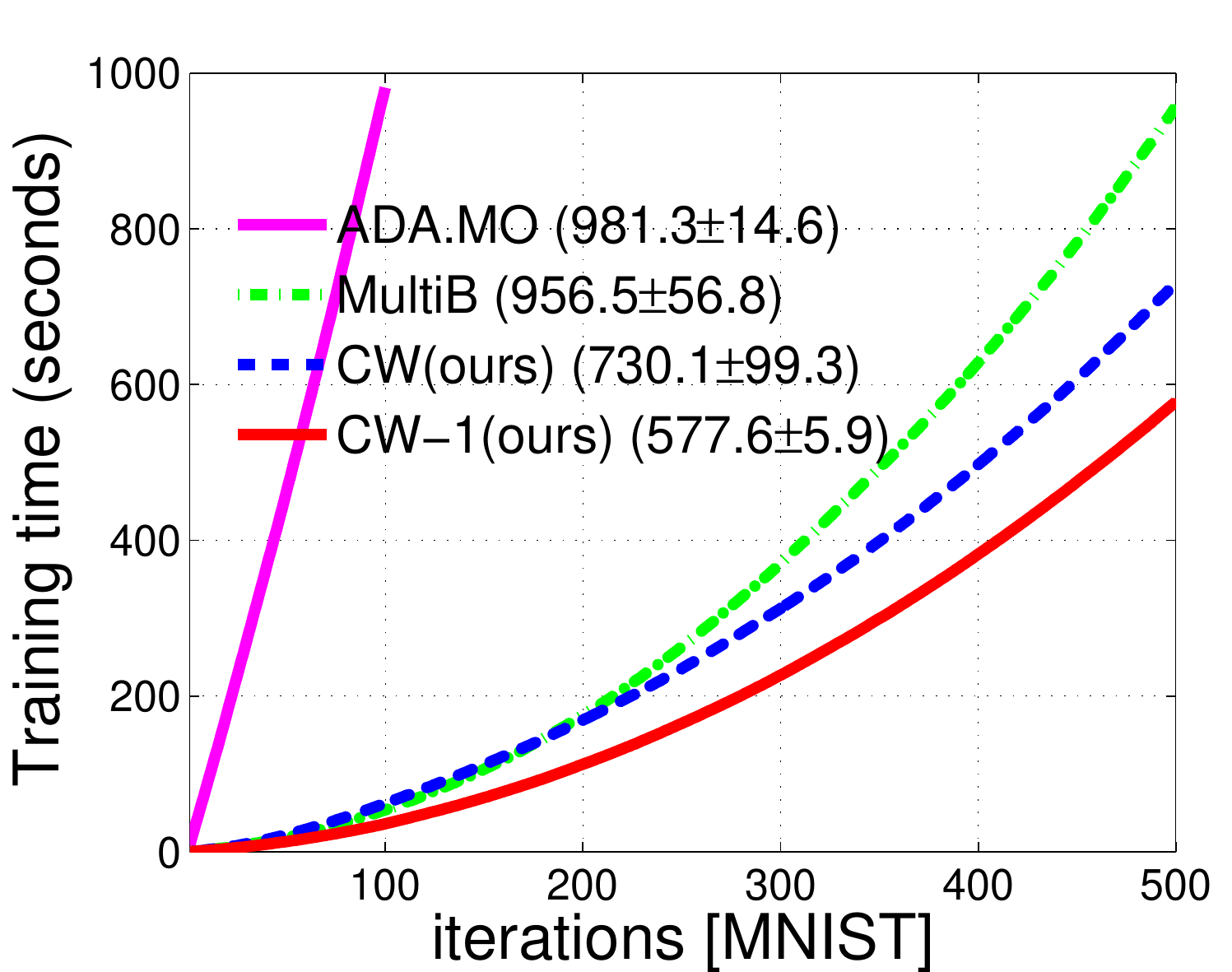}
   	    	    \includegraphics[width=.33\linewidth]{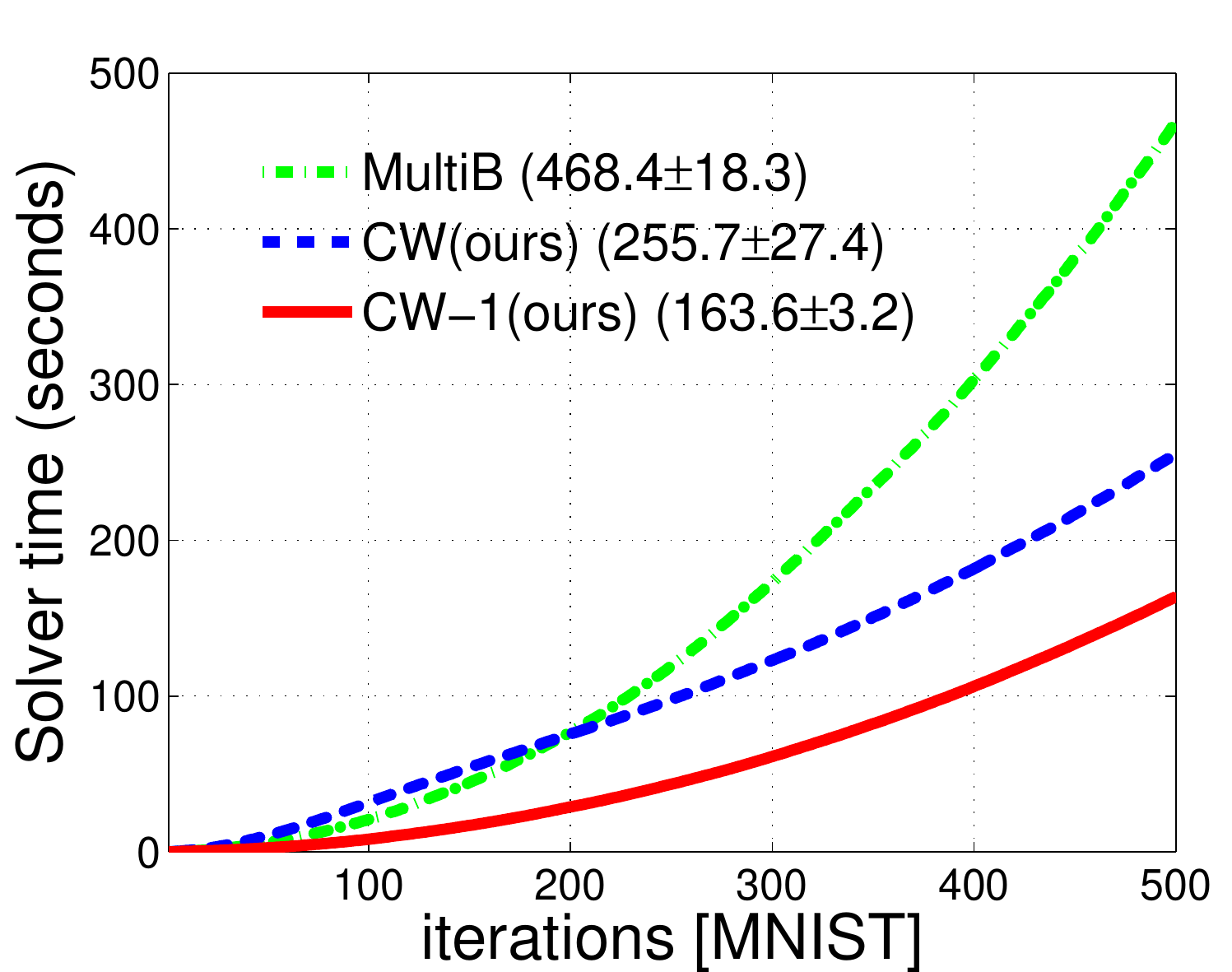}
    }
    \caption{Experiments on 3 handwritten digit recognition datasets: USPS,
    PENDIGITS and MNIST. CW and CW-1 are our methods. CW-1 uses
    stage-wise setting. Our methods converge much faster, achieve best
    test error and use less training time. Ada.MO has similar
    convergence rate as ours, but requires much more training time.
    With a maximum training time of 1000 seconds, Ada.MO failed to
    finish 500 iterations on all 3 datasets.}
    \label{fig:hand}
\end{figure}

AdaBoost.MO \cite{Schapire99improvedboosting} (Ada.MO) has a similar
convergence rate as our method, but it is much slower than our method
and becomes intractable for large scale datasets. We run Ada.MO on some UCI
datasets and MNIST. Results are shown in Fig. \ref{fig:uci} and
Fig. \ref{fig:hand}. We set a maximum training time (1000 seconds) for
Ada.MO; other methods are all below this maximum time on those
datasets. If maximum time reached, we report the results of those
finished iterations.

{\bf UCI datasets}: we use 2 UCI multi-class datasets: VOWEL and
ISOLET. For each dataset, we randomly select 75\% data for training
and the rest for testing. Results are shown in Fig. \ref{fig:uci}.

{\bf Handwritten digit recognition}: we use 3 handwritten datasets:
MNIST, USPS and PENDIGITS. For MNIST, we randomly sample 1000 examples
from each class, and use the original test set of 10,000 examples. For
USPS and PENDIGITS, we randomly select 75\% for training, the rest for
testing. Results are shown in Fig. \ref{fig:hand}.

\begin{figure}[h!]
    \centering

	    \includegraphics[width=.42\linewidth]{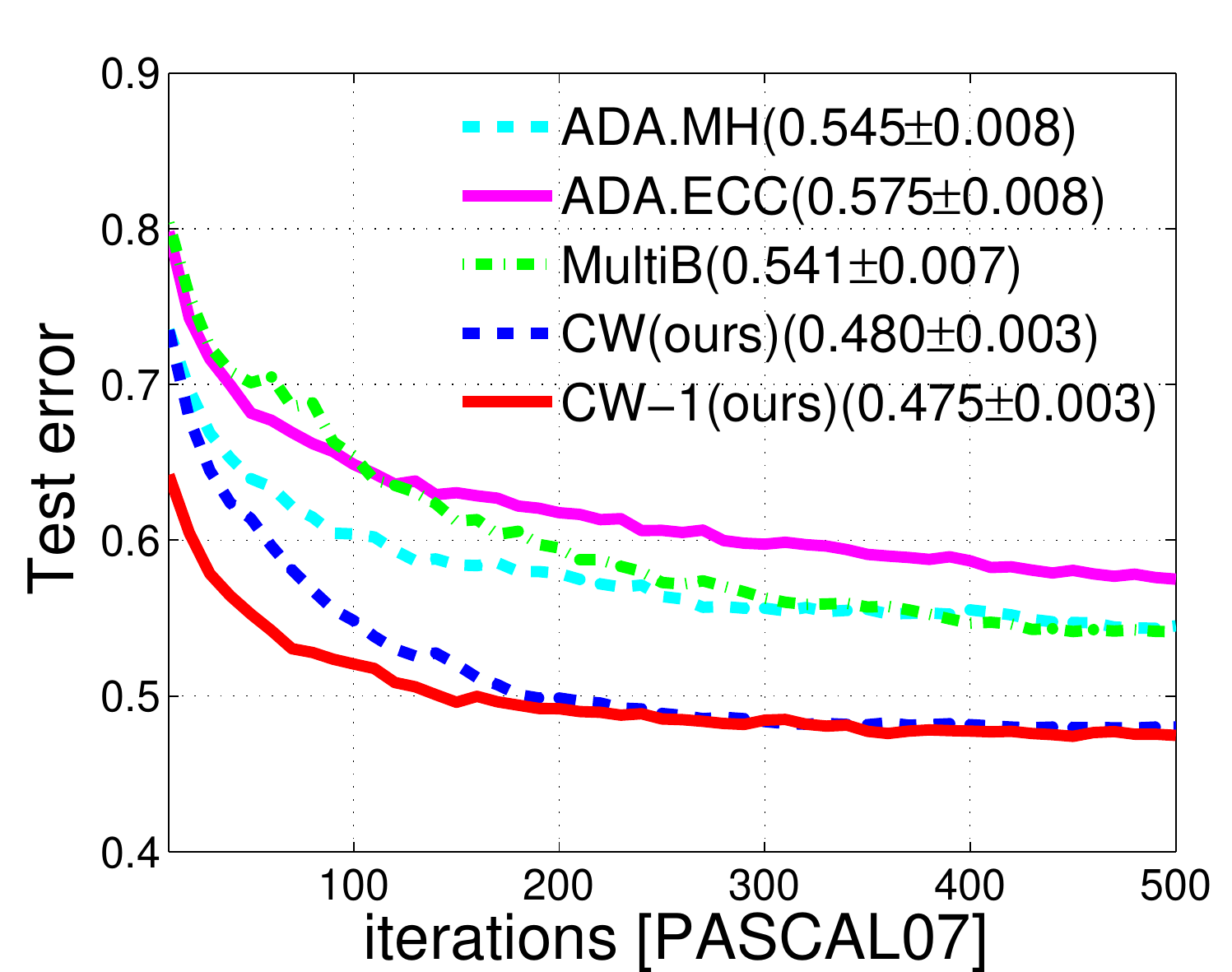}
    	    \includegraphics[width=.42\linewidth]{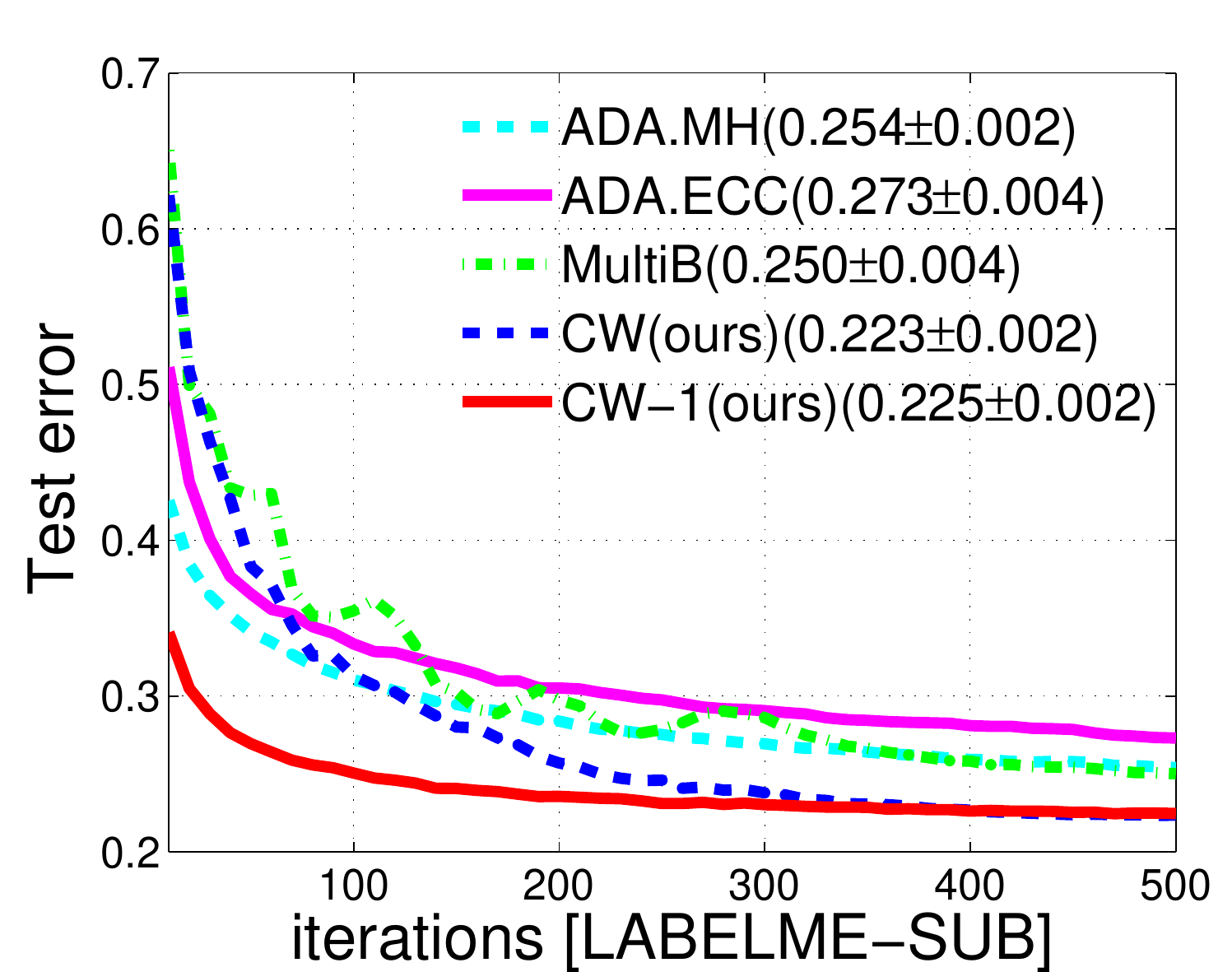}
	    \includegraphics[width=.42\linewidth]{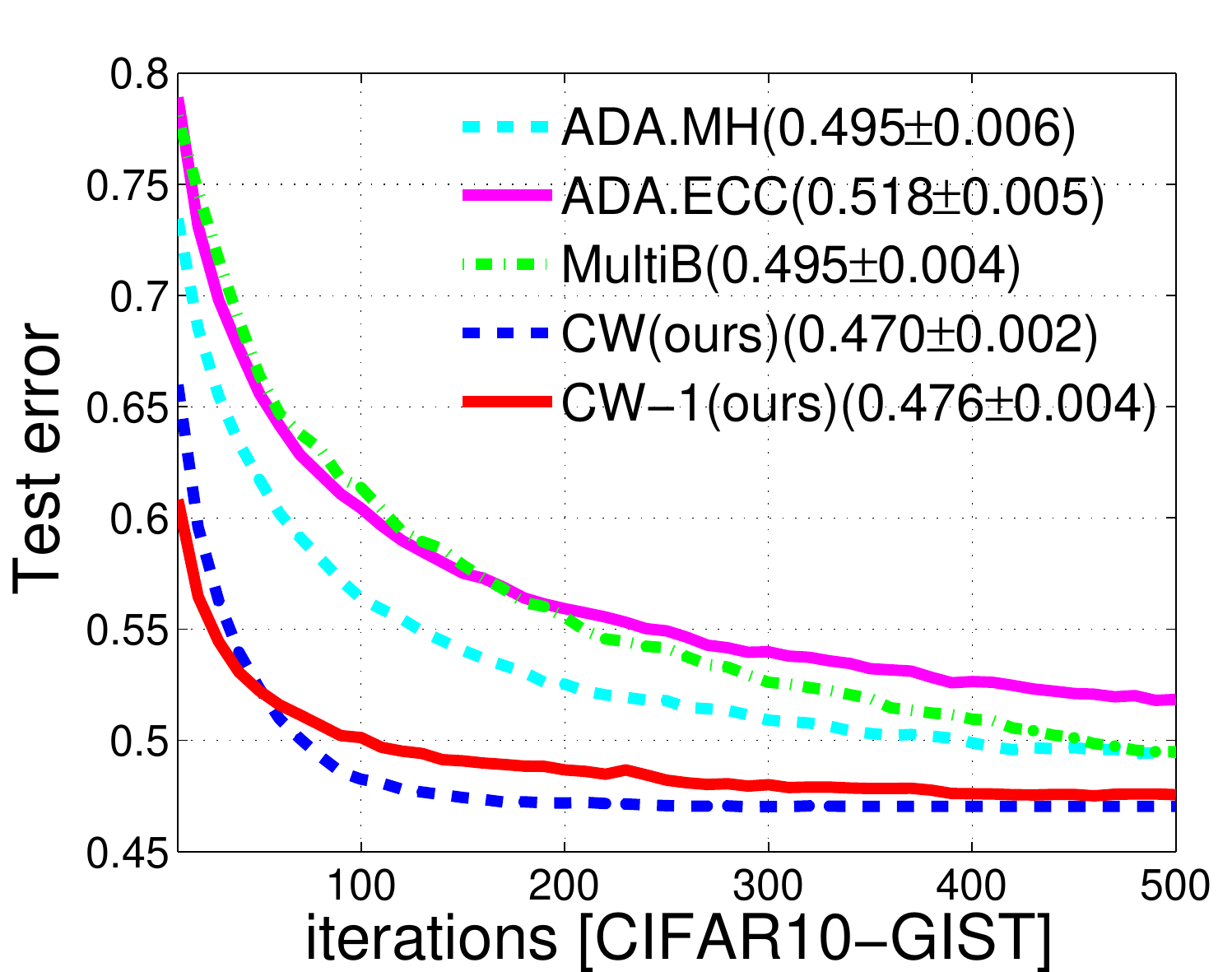}
	    \includegraphics[width=.42\linewidth]{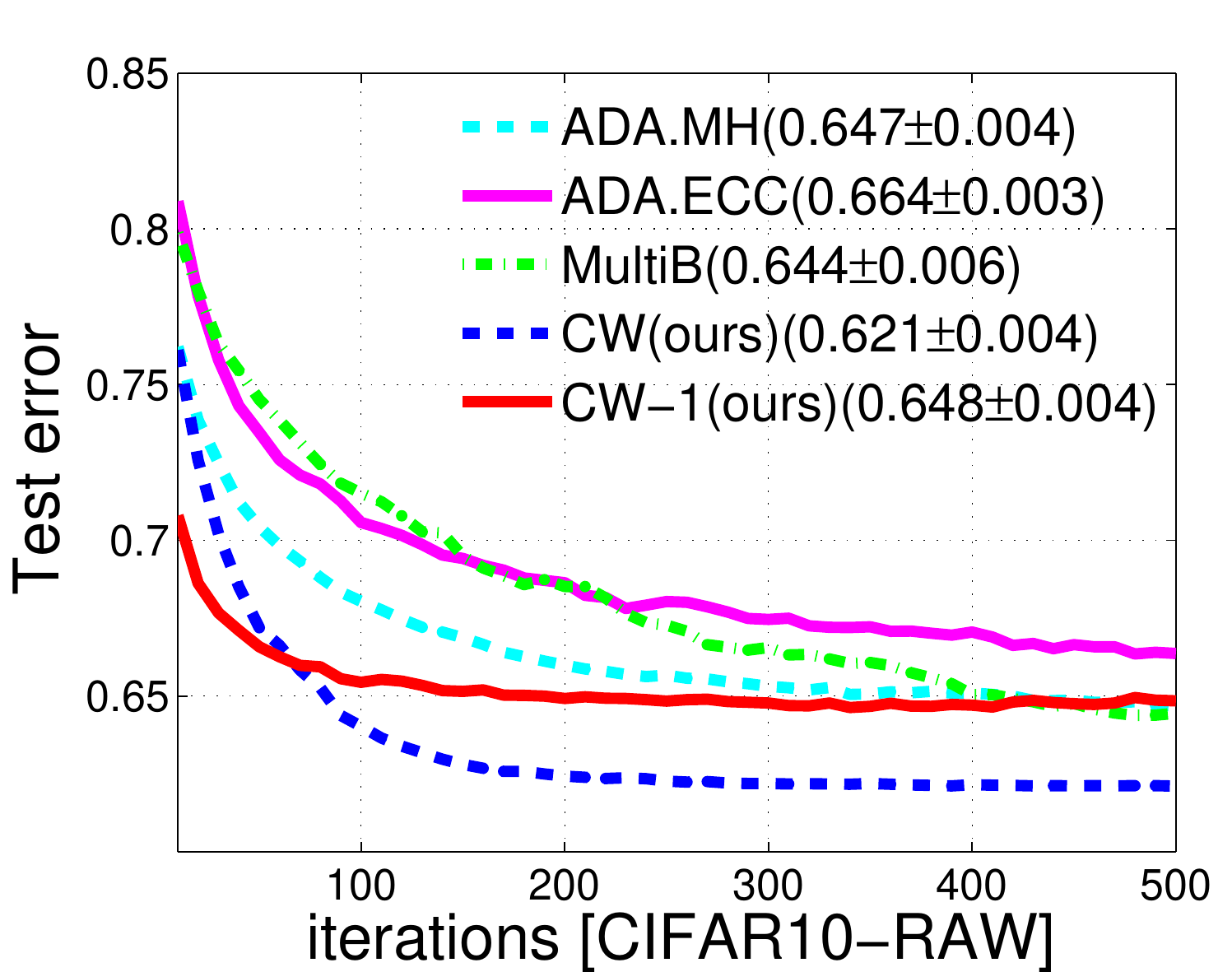}
	    \includegraphics[width=.42\linewidth]{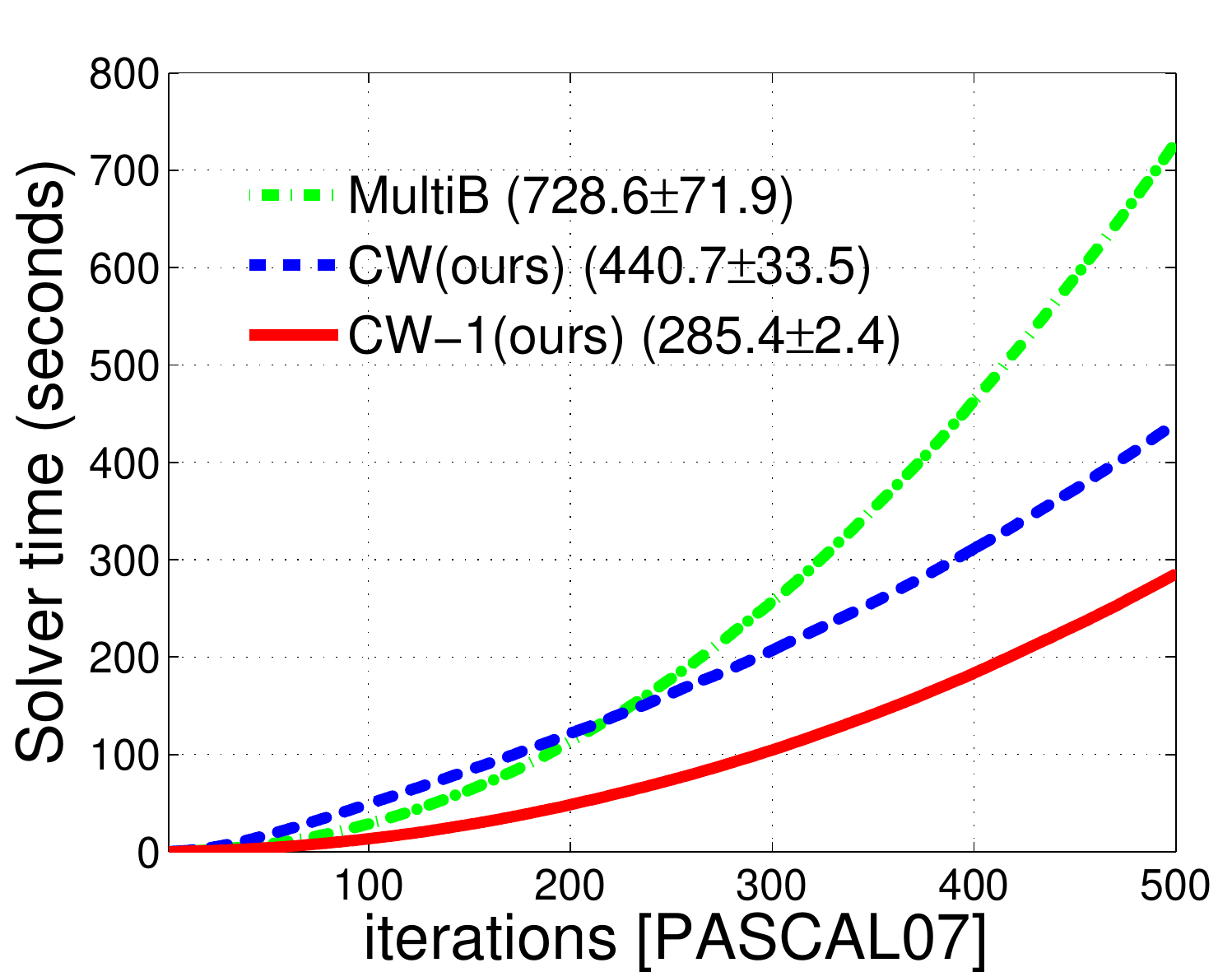}
	    \includegraphics[width=.42\linewidth]{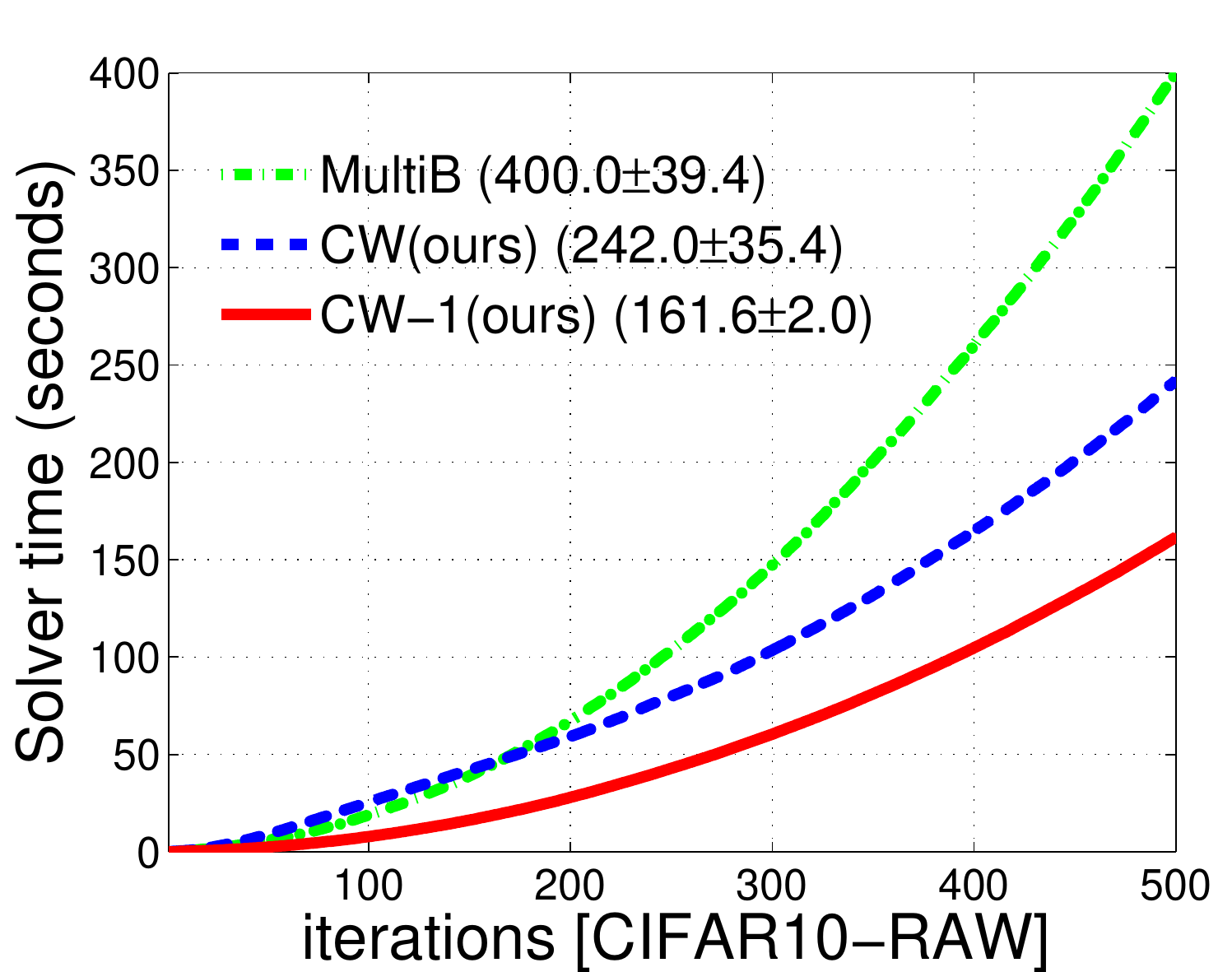}
	    \includegraphics[width=.42\linewidth]{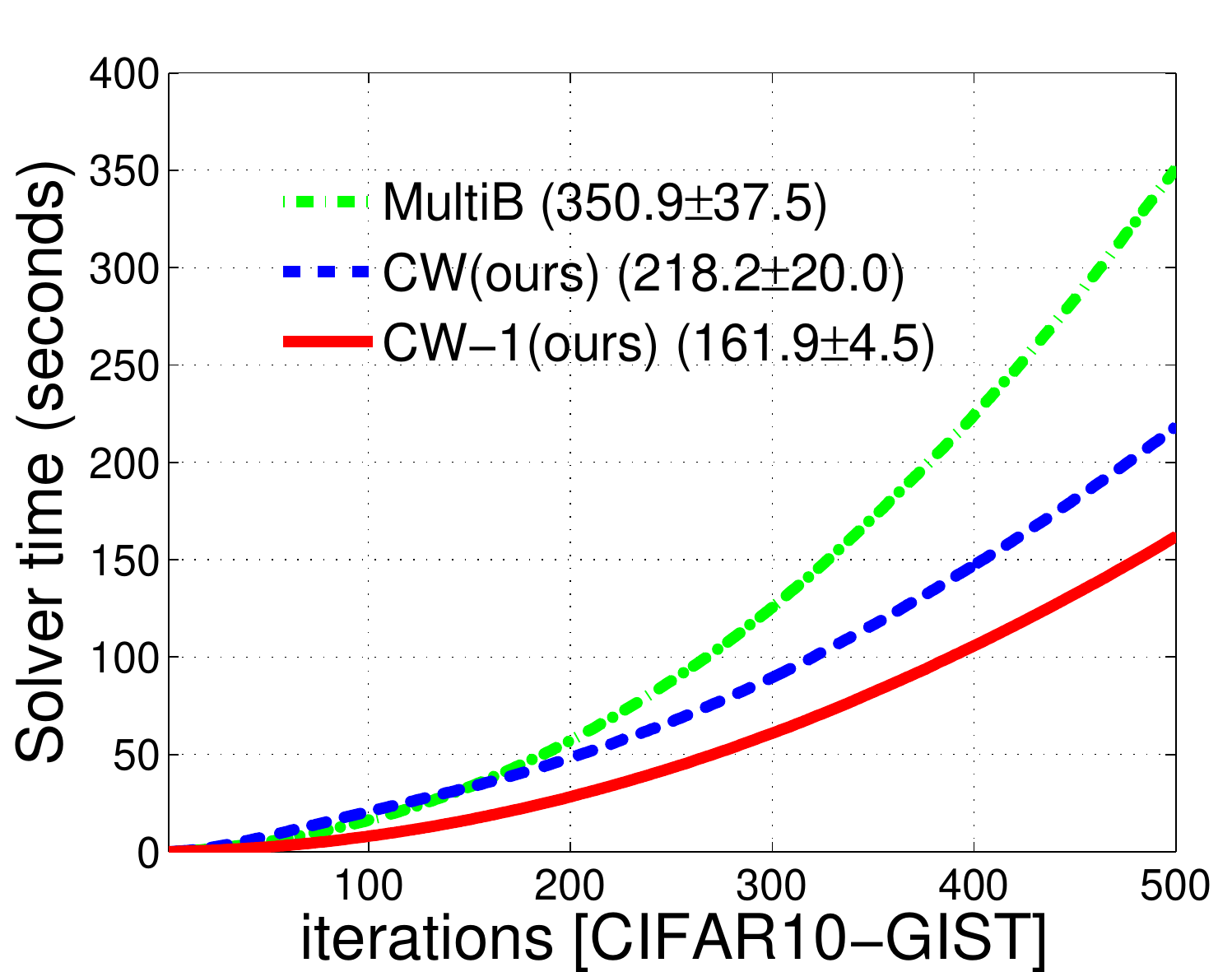}

    \caption{Experiments on 3 image datasets: PASCAL07, LabelMe and CIFAR10. CW and CW-1 are our methods. CW-1 uses stage-wise setting.
Our methods converge much faster, achieve best test error and use less training time.}
    \label{fig:images}
\end{figure}

{\bf 3 Image datasets: PASCAL07, LabelMe, CIFAR10}:
For PASCAL07, we use 5 types of features provided in \cite{guillaumin2010}.
For labelMe, we use the subset:
LabelMe-12-50k\footnote{http://www.ais.uni-bonn.de/download/datasets.html}
and generate GIST features. For these two datasets, we use those
images which only have one class label. We use 70\% data for training,
the rest for testing. For
CIFAR10\footnote{http://www.cs.toronto.edu/\~{}kriz/cifar.html},
we construct 2 datasets, one uses GIST features and
the other uses the pixel values. We use the provided test set and 5
training sets for 5 times run. Results are shown in
Fig. \ref{fig:images}.

\begin{figure}[t]
    \centering

   \subfloat{
    	    \includegraphics[width=.33\linewidth]{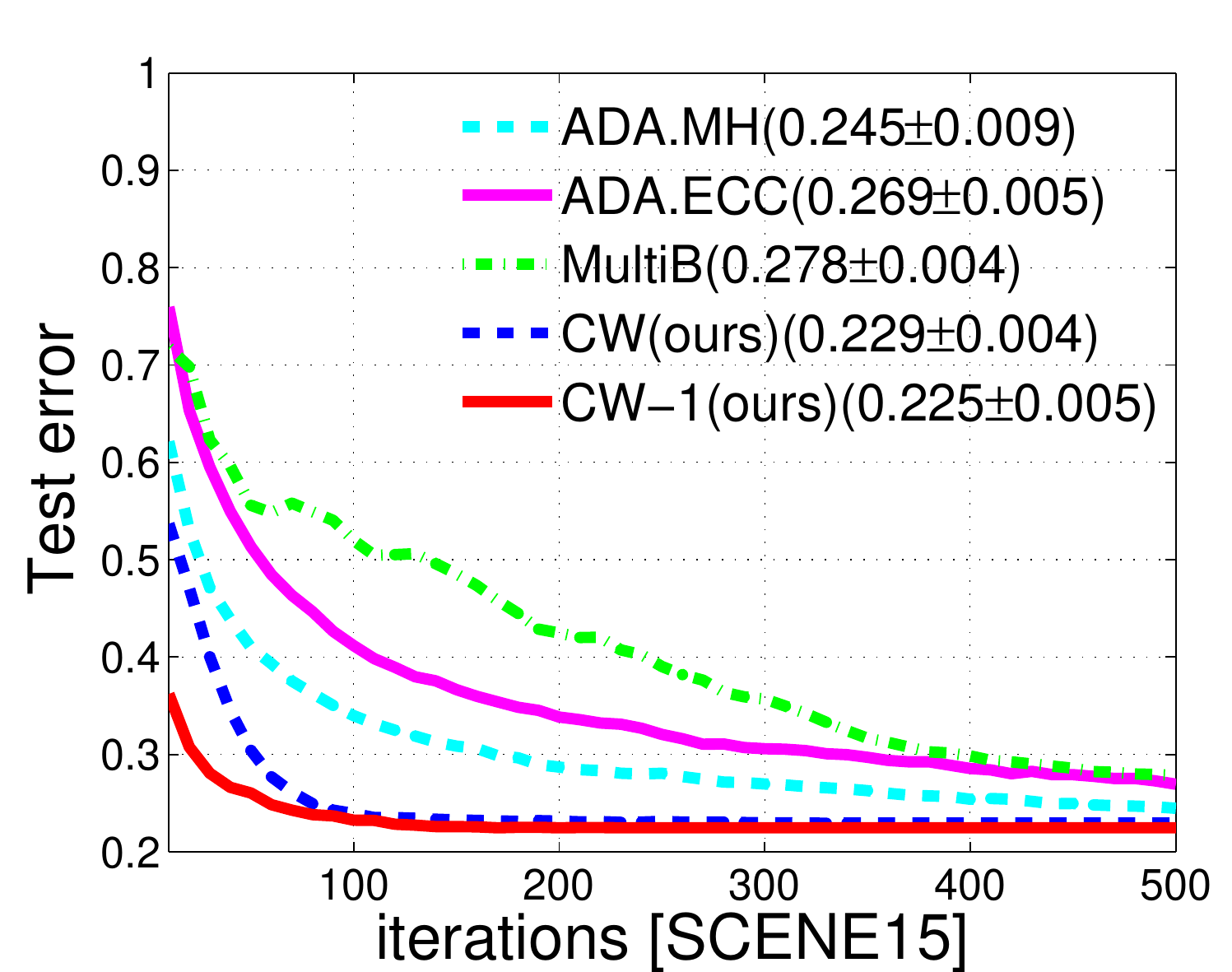}
    	    \includegraphics[width=.33\linewidth]{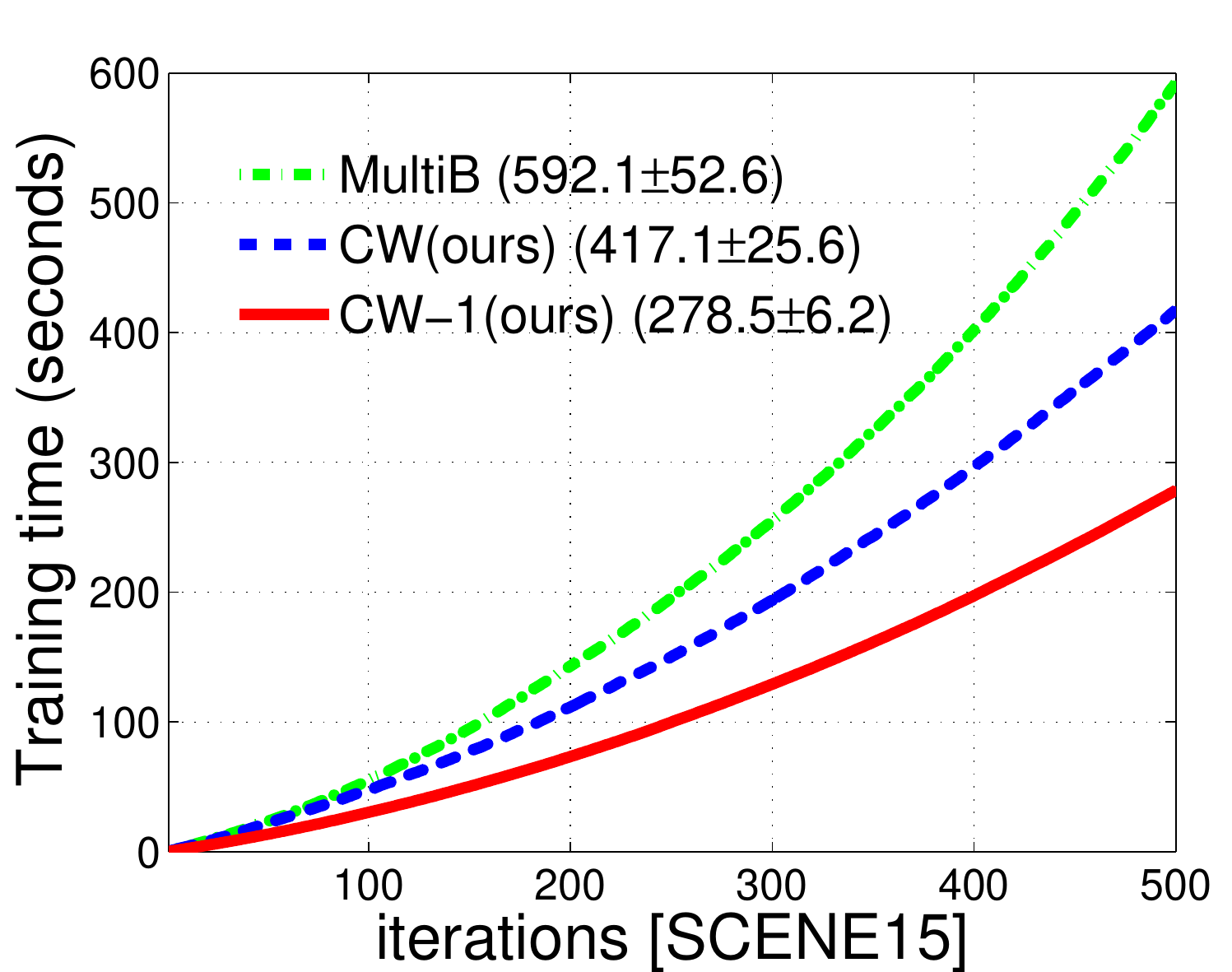}
    	    \includegraphics[width=.33\linewidth]{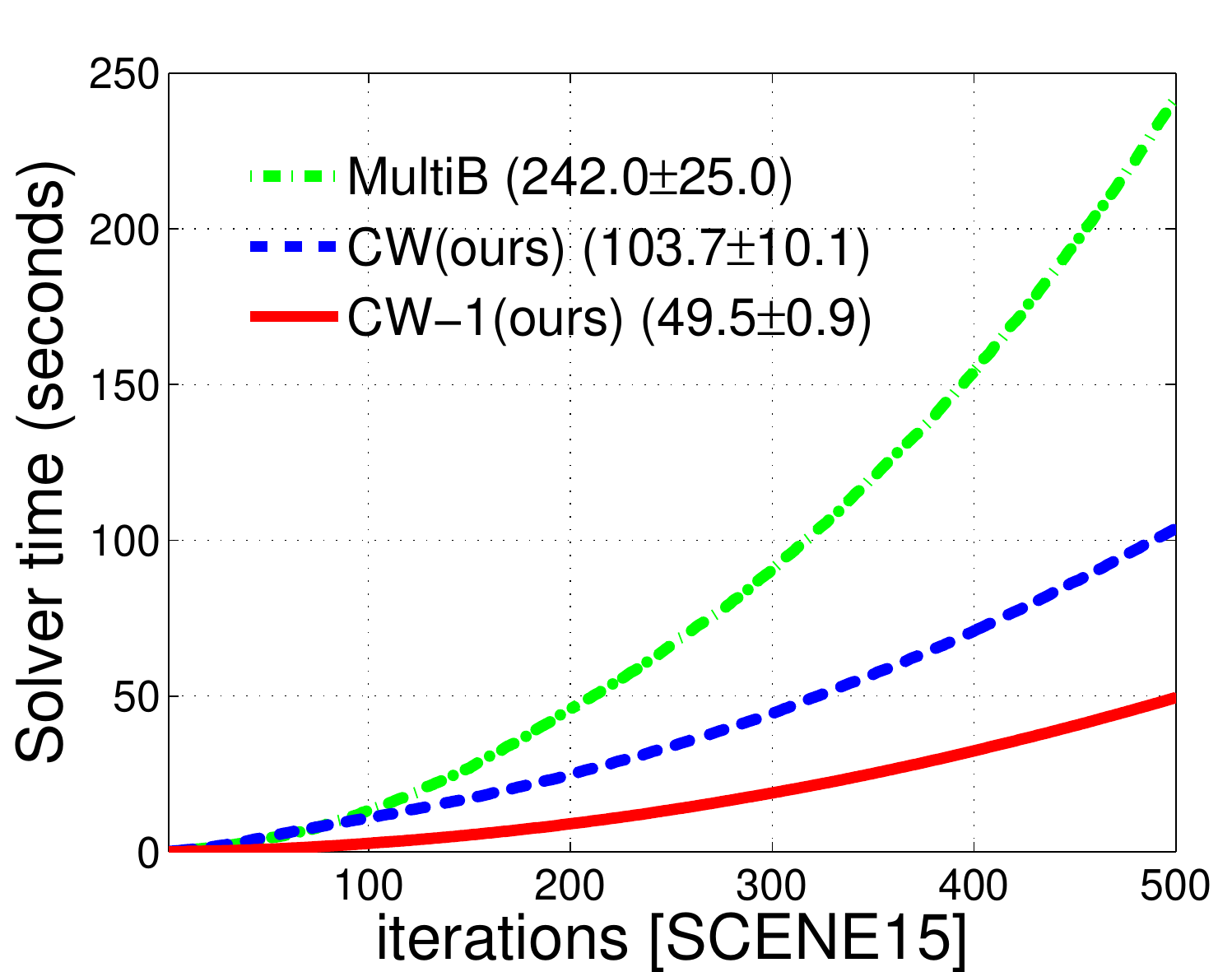}
    }

        \subfloat{
    	    \includegraphics[width=.33\linewidth]{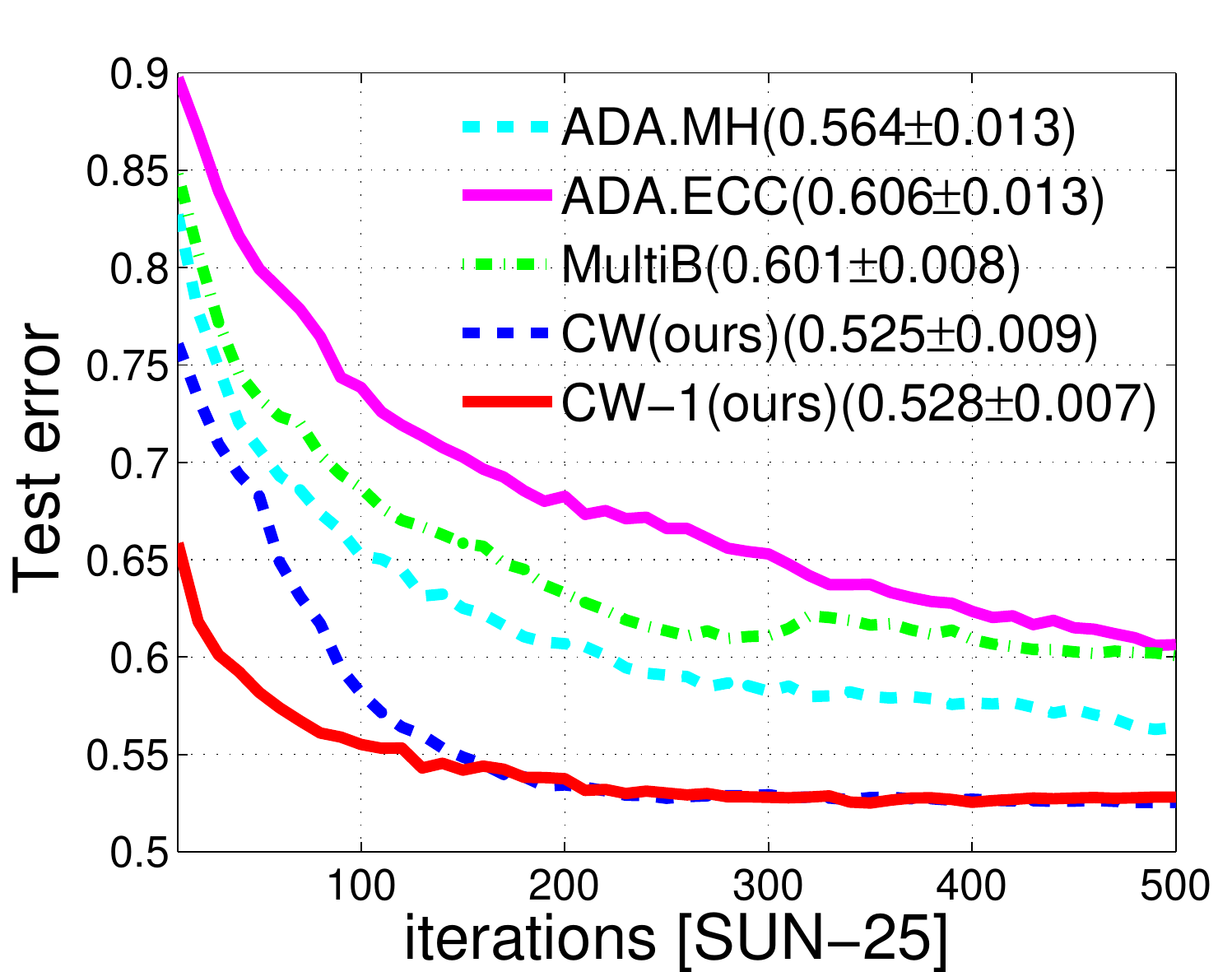}
    	    \includegraphics[width=.33\linewidth]{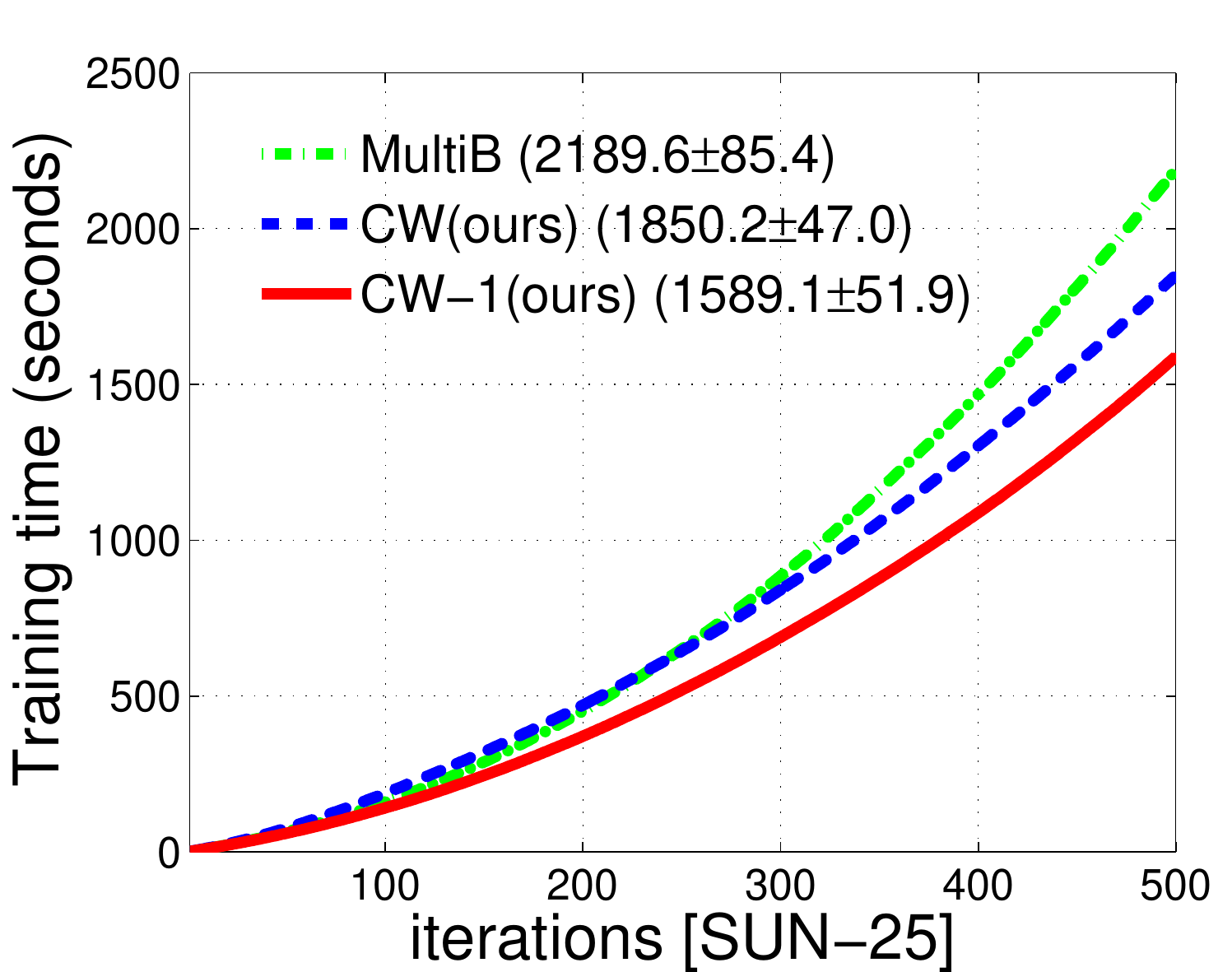}
    	    \includegraphics[width=.33\linewidth]{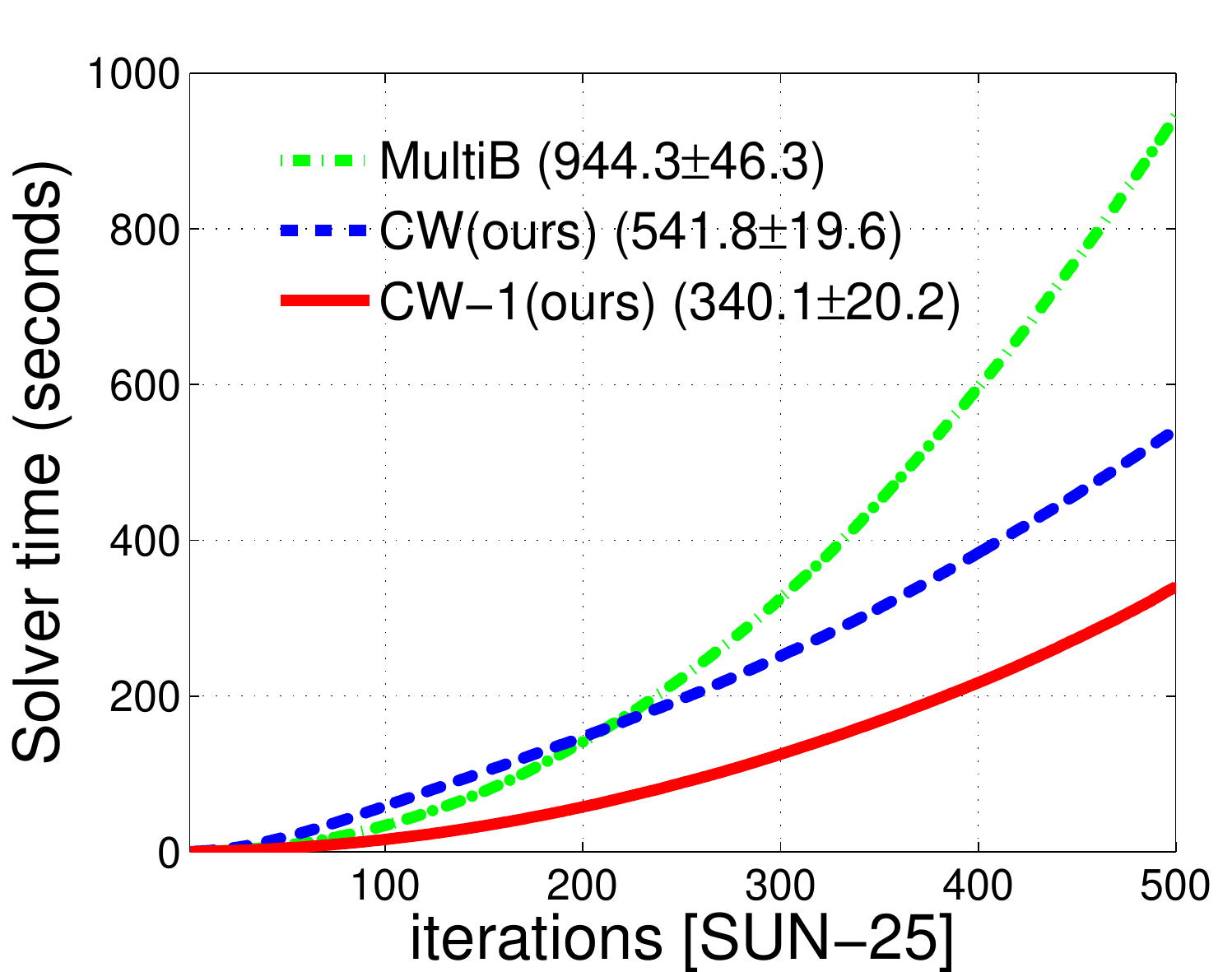}
    }

    \caption{Experiments on 2 scene recognition datasets: SCENE15 and
    a subset of SUN. CW and CW-1 are our methods. CW-1 uses stage-wise
    setting. Our methods converge much faster and achieve best test
    error and use less training time.}
    \label{fig:scene}
\end{figure}

{\bf Scene recognition}: we use 2 scene image datasets:
Scene15 \cite{lazebnik2006} and SUN \cite{xiao2010}. For Scene15, we
randomly select 100 images per class for training, and the rest for
testing. We generate histograms of code words as features. The code
book size is 200. An image is divided into 31 sub-windows in a spatial
hierarchy manner. We generate histograms in each sub-windows, so the
histogram feature dimension is 6200. For SUN dataset, we construct a
subset of the original dataset containing 25 categories. For each
category, we use the top 200 images, and randomly select 80\% data for
training, the rest for testing. We use the HOG features described
in\cite{xiao2010}. Results are shown in Fig. \ref{fig:scene}.

\begin{figure}[t]
    \centering
    \subfloat{
	    \includegraphics[width=.443\linewidth]{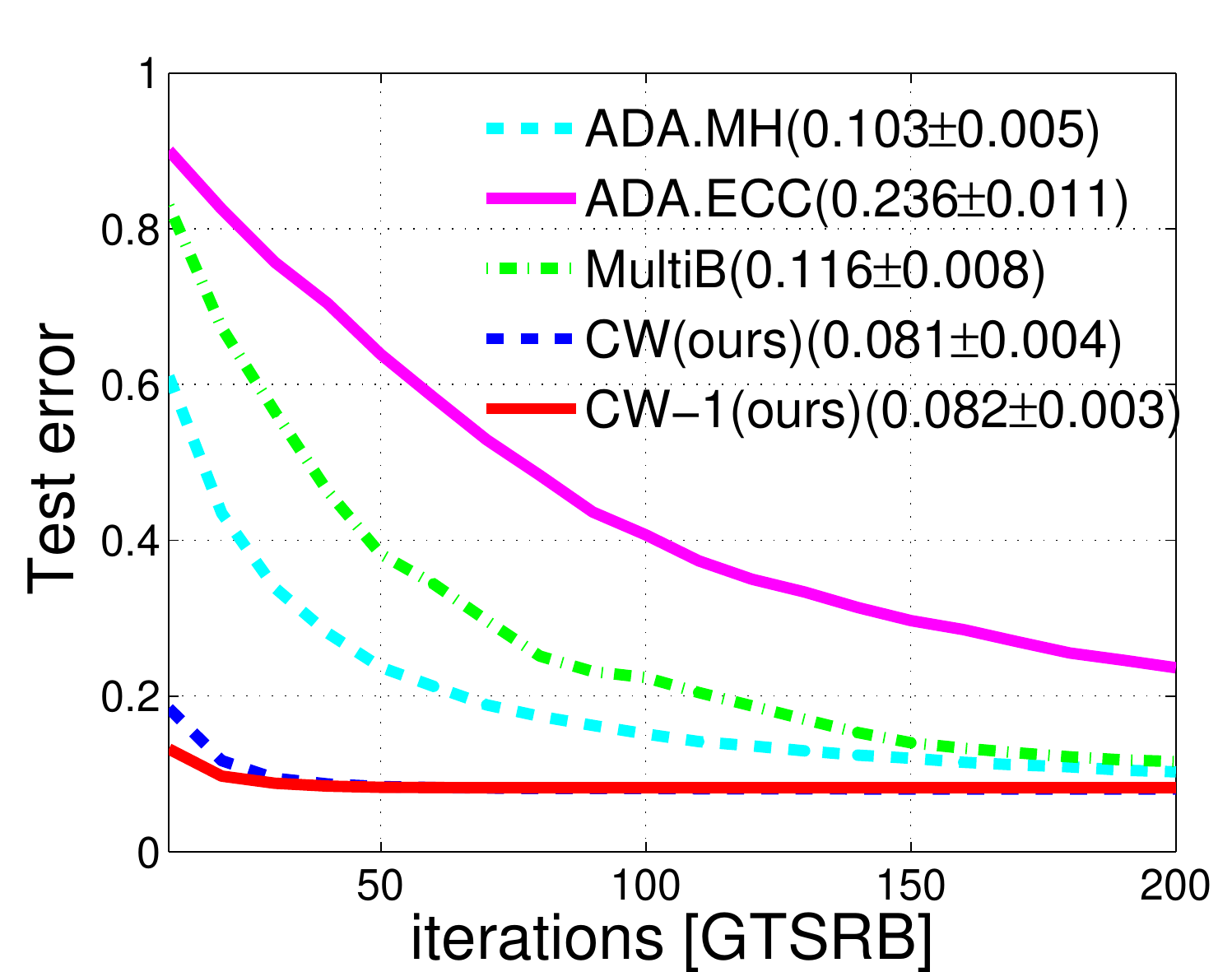}
   	    \includegraphics[width=.443\linewidth]{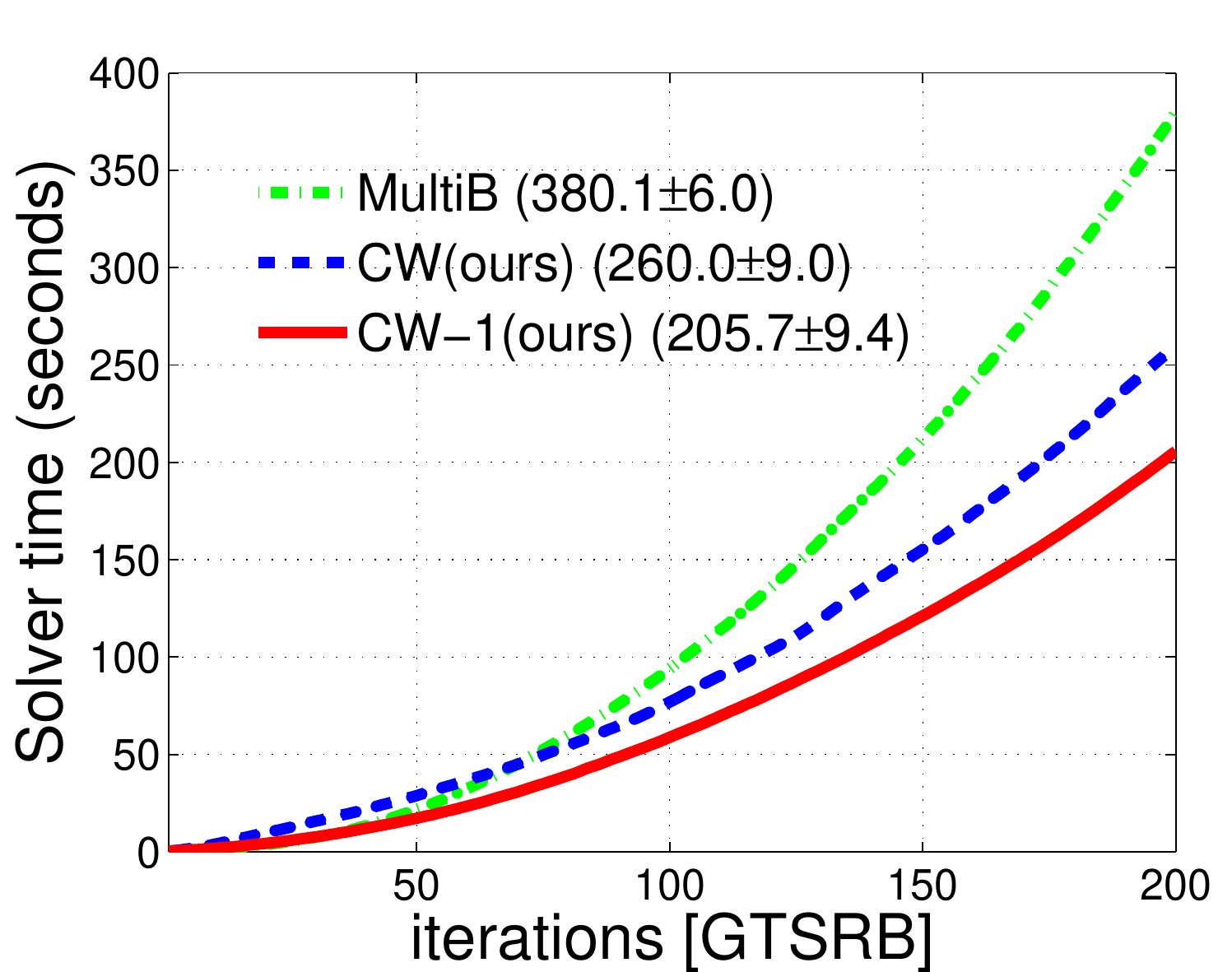}
    }

    \caption{Results on the traffic sign dataset: GTSRB.
    CW and CW-1 (stage-wise setting) are our methods. Our
    methods converge much faster, achieve best test error and use less
    training time.}
    \label{fig:traffic}
\end{figure}

{\bf Traffic sign recognition}: We use the GTSRB\footnote{http://benchmark.ini.rub.de/}
 traffic sign dataset. There are 43 classes and more than 50000
 images. We use the provided 3 types of HOG features; so there are
 6052 features in total. We randomly select 100 examples per class for
 training and use the original test set. Results are shown in
 Fig.\ref{fig:traffic}.

\subsection{\scd evaluation}

\begin{figure}[t]
    \centering

    \subfloat{
    	    \includegraphics[width=.33\linewidth]{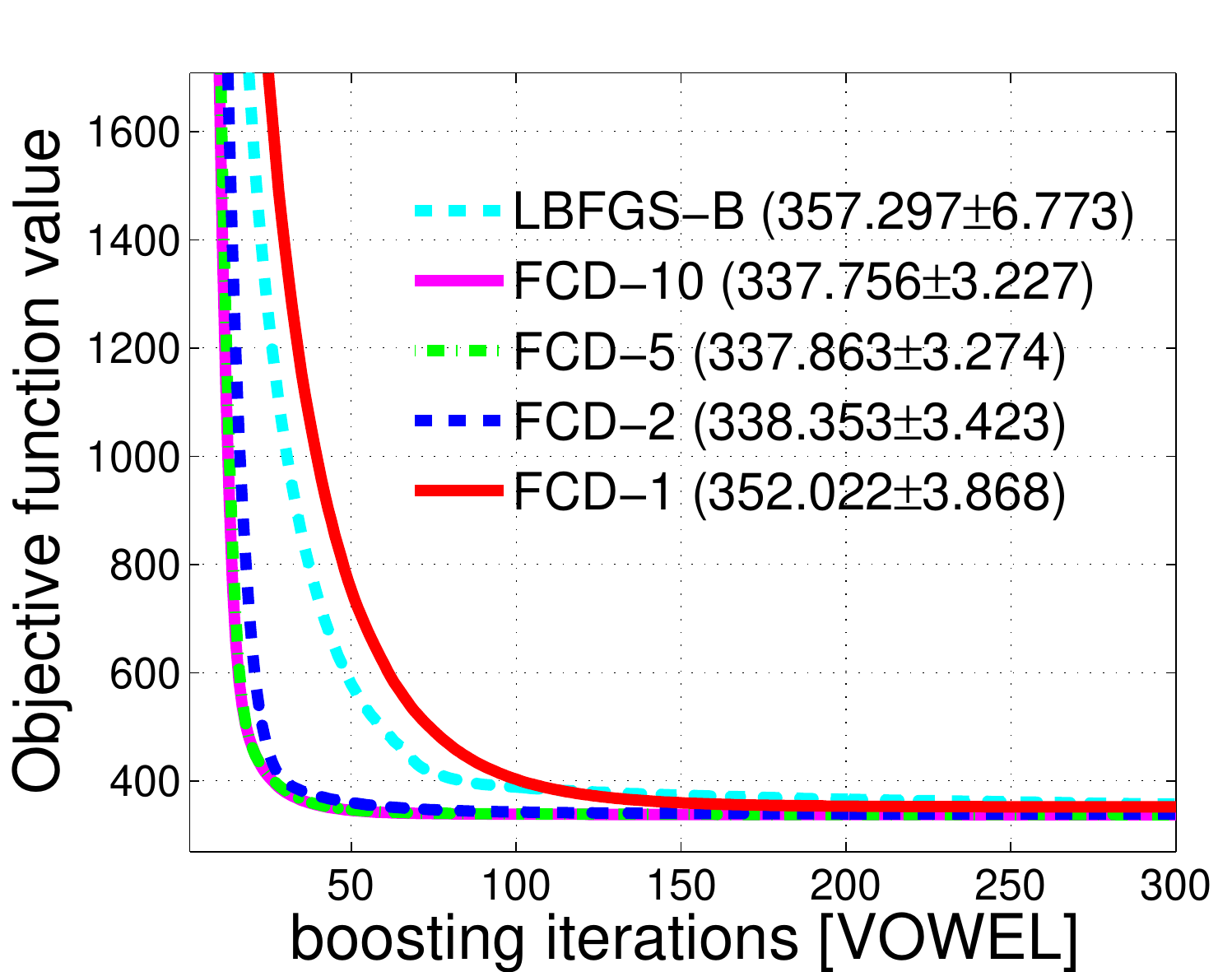}
    	    \includegraphics[width=.33\linewidth]{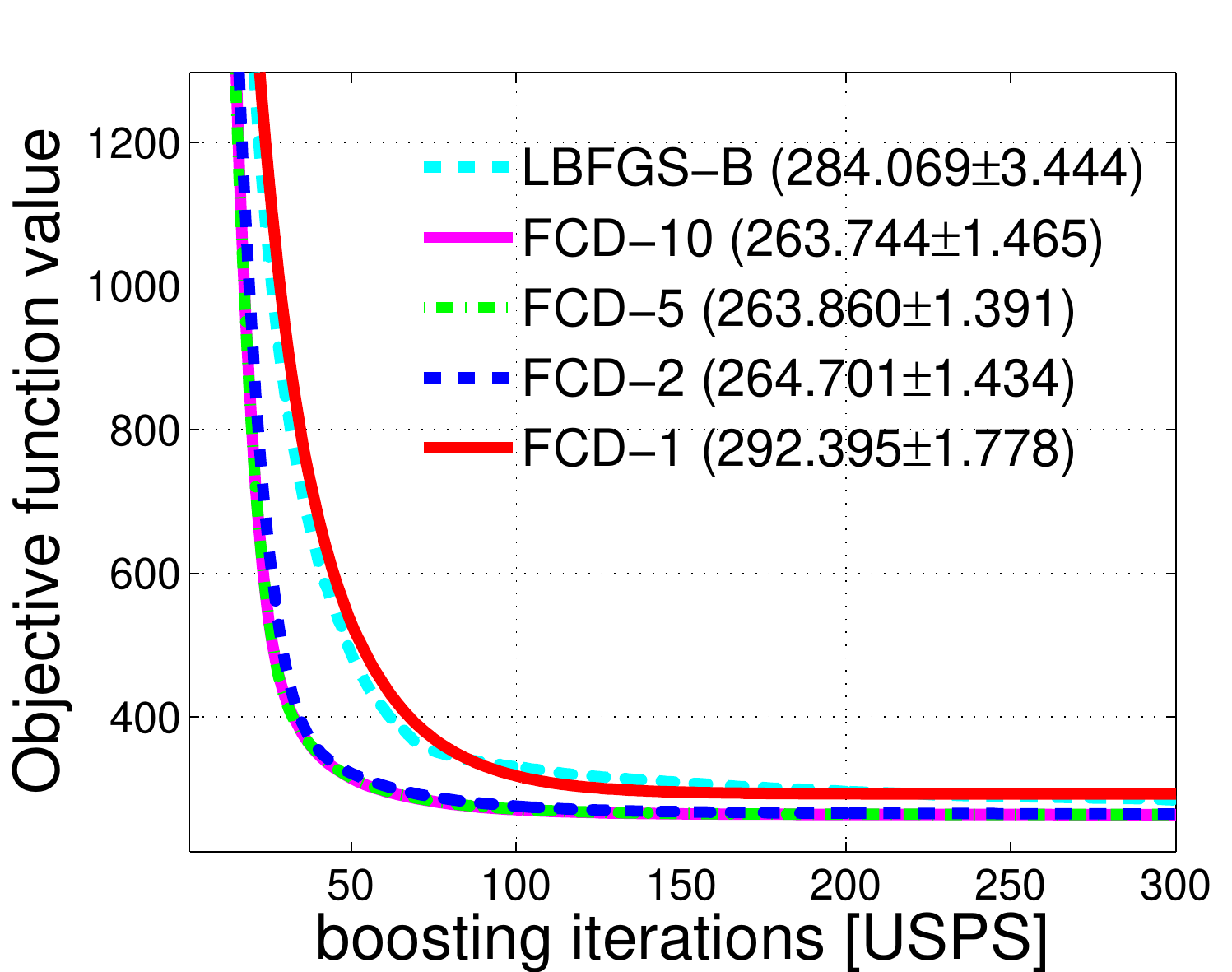}
       	    \includegraphics[width=.33\linewidth]{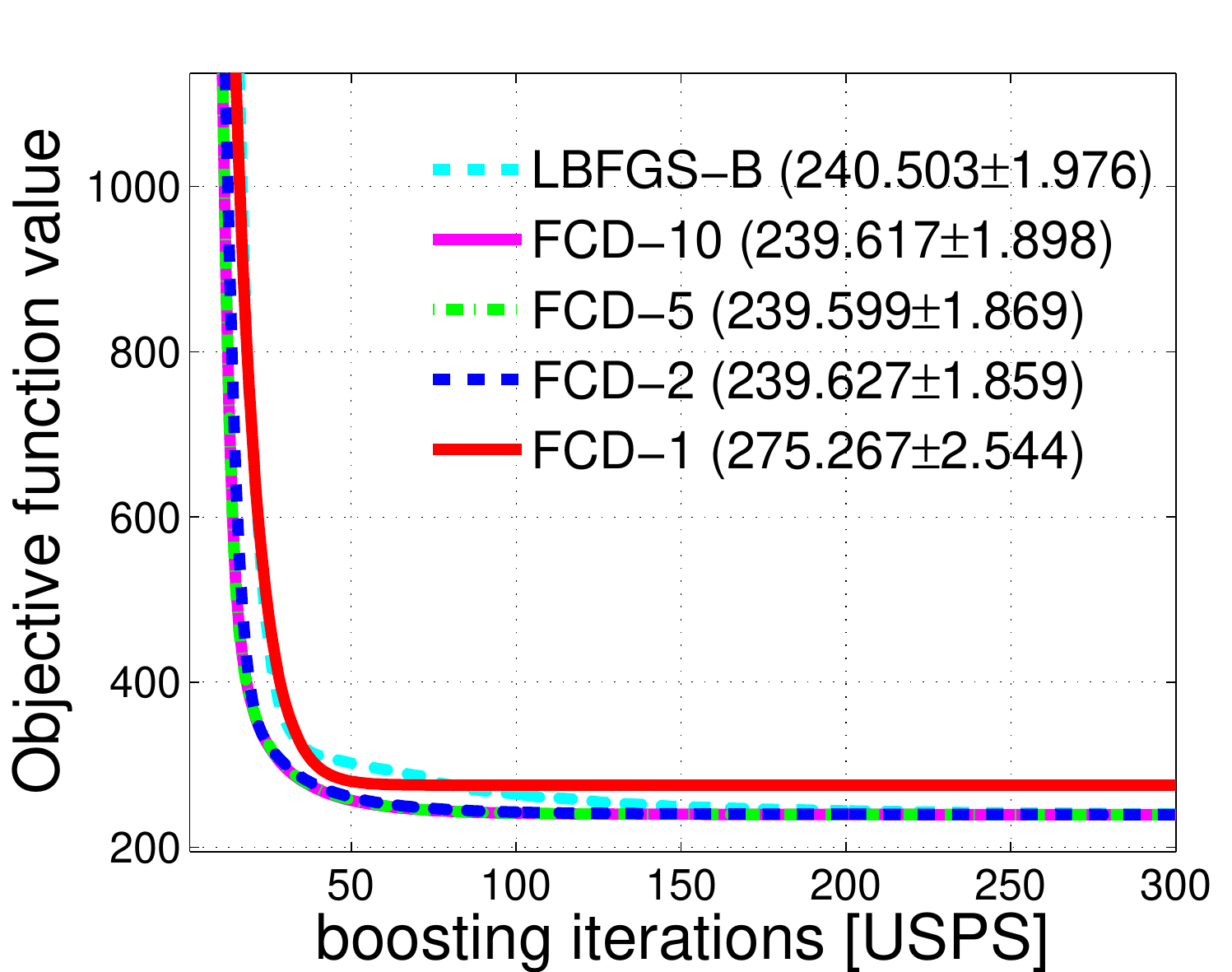}

    }

     \subfloat{
    	    \includegraphics[width=.33\linewidth]{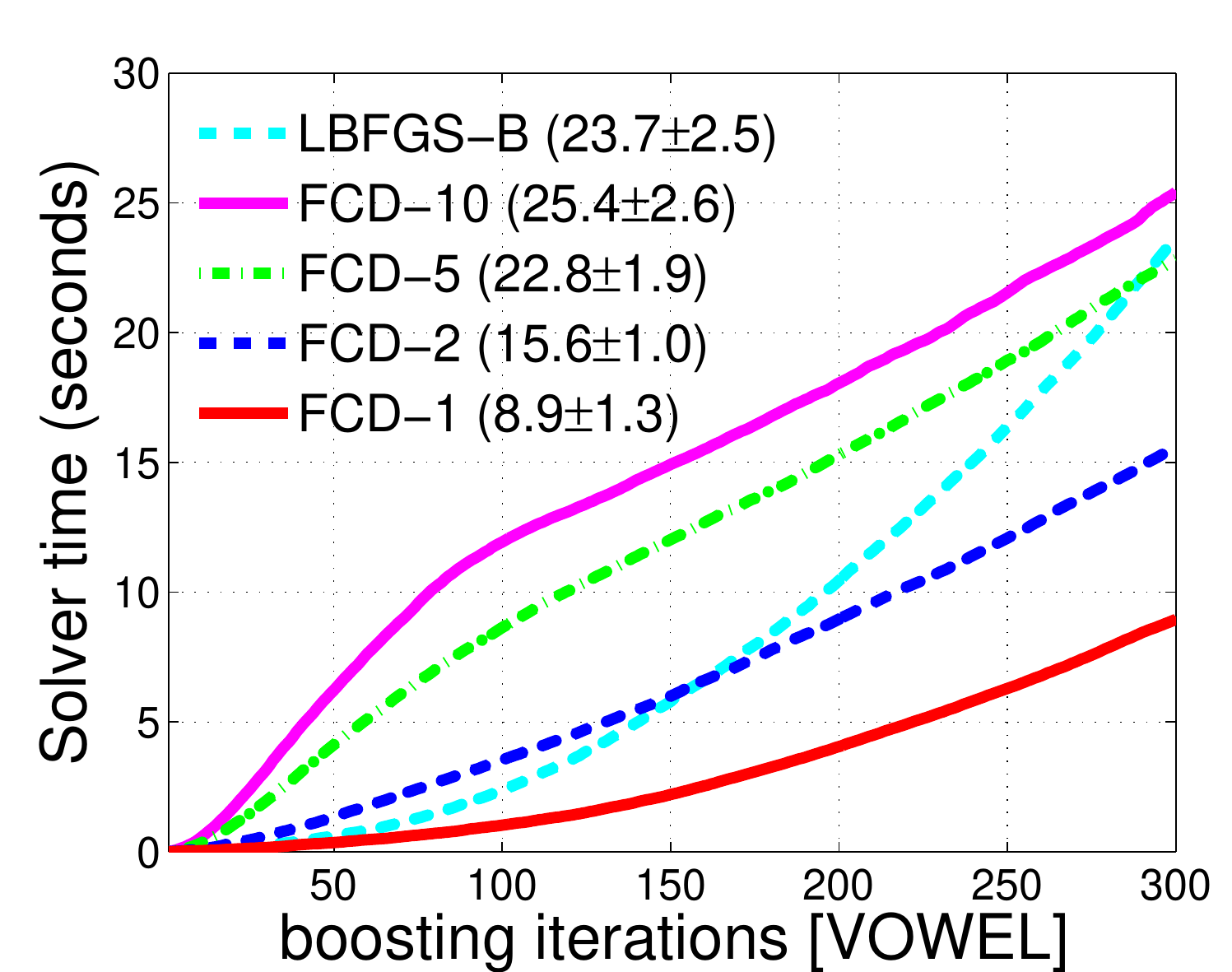}
    	    \includegraphics[width=.33\linewidth]{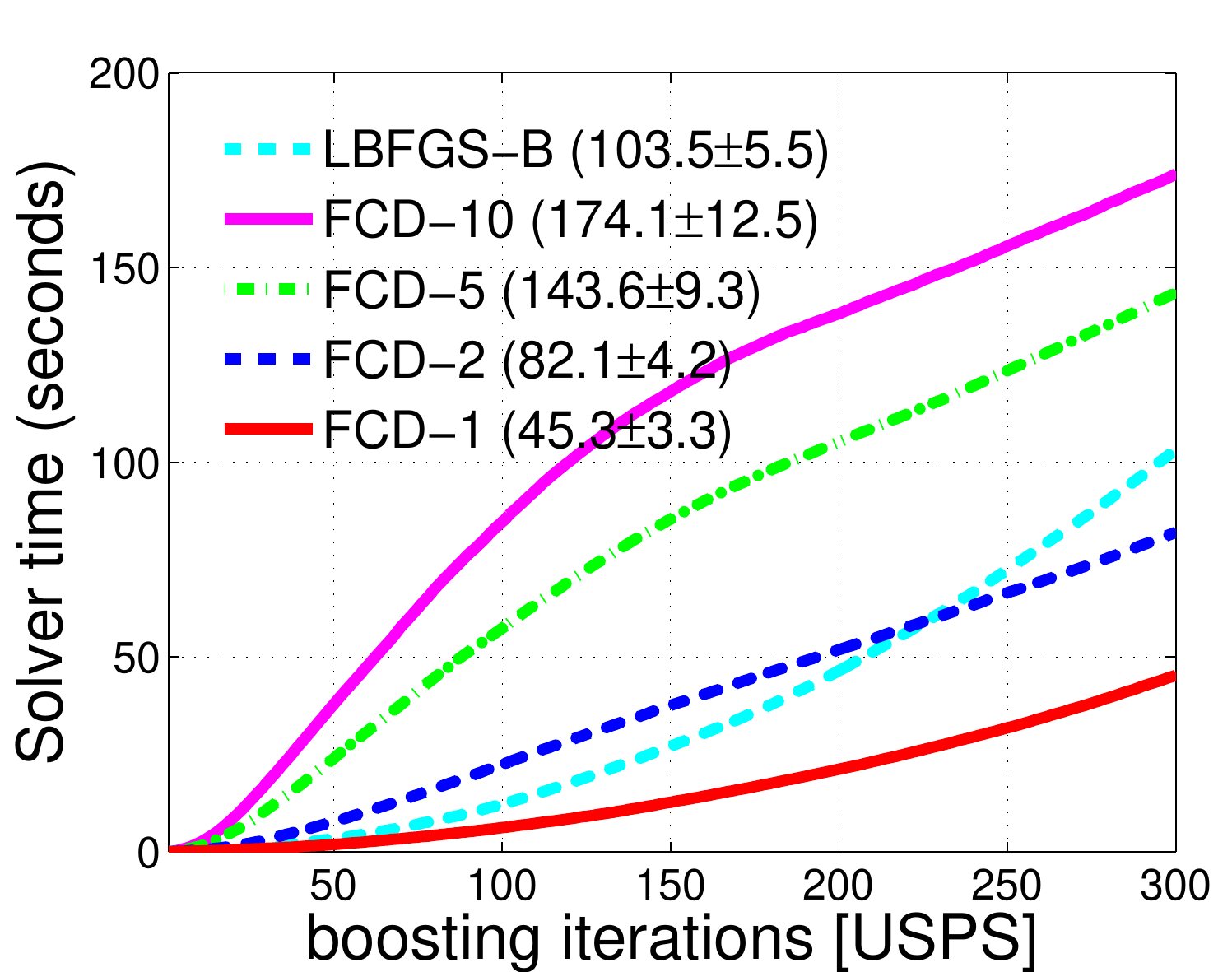}
       	    \includegraphics[width=.33\linewidth]{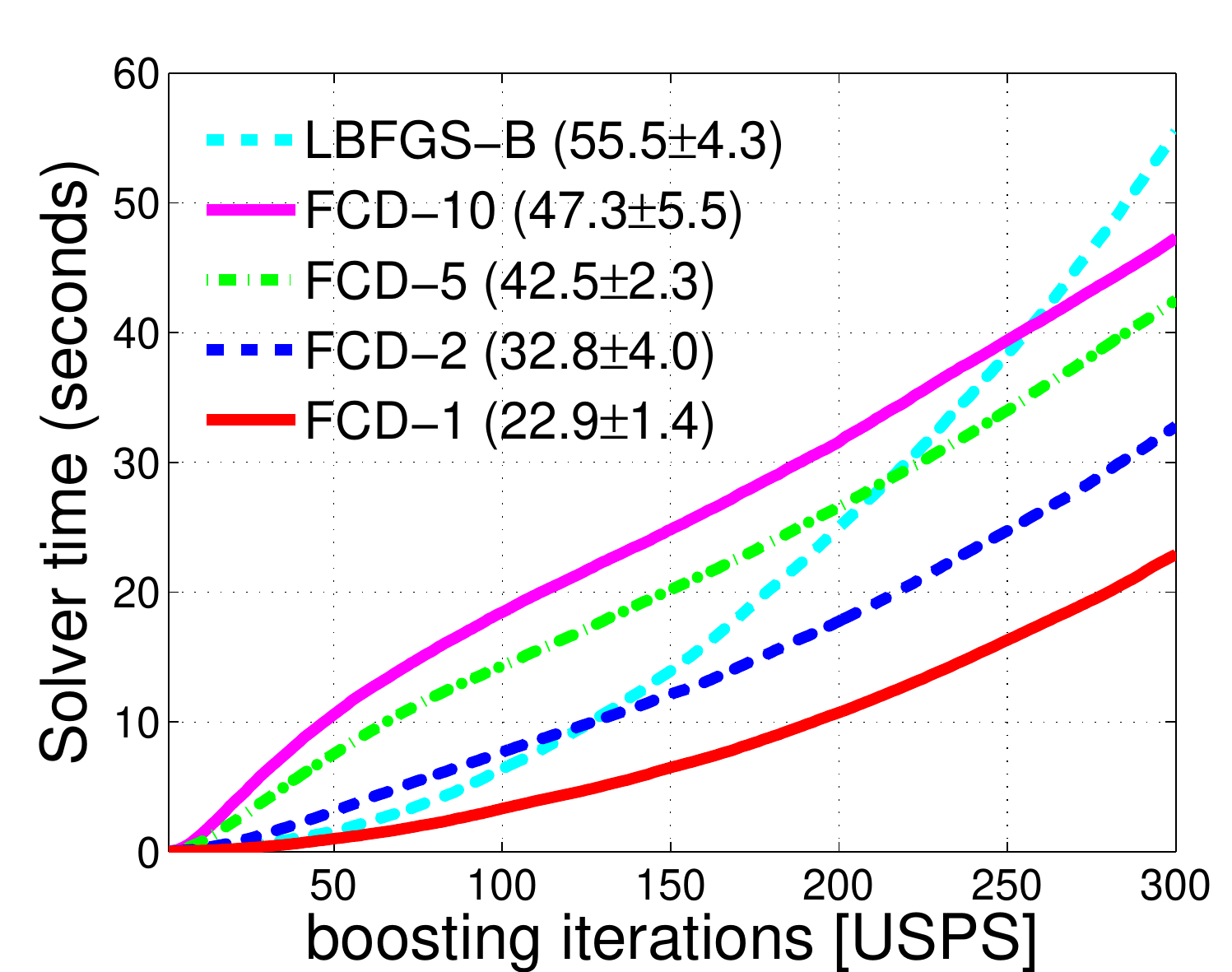}

    }

    \caption{Solver comparison between \scd with different parameter
    setting and LBFGS-B \cite{lbfgs}. One column for one dataset.
    The number after ``FCD'' is the setting for the maximum iteration ($\Z$)
    of \scd. The stage-wise setting of \scd is the fastest one.
    See the text for details.
    }
    \label{fig:fcd}
\end{figure}

We perform further experiments to evaluate \scd with different
parameter settings, and compare to the LBFGS-B \cite{lbfgs} solver.
We use 3
datasets in this section: VOWEL, USPS and SCENE15. We run \scd with
different settings of the maximum working set iteration($\Z$ in Algorithm \ref{ALG:alg2}) to evaluate how the
setting of $\Z$ (maximum working set iteration) affects the performance of \scd. We also run LBFGS-B
\cite{lbfgs} solver for solving the same optimization \eqref{eq:mcgcon} as \scd.
We set $C=10^4$ for all cases.
Results are shown in Fig. \ref{fig:fcd}. For LBFGS-B, we use the default converge setting to get a moderate solution.
The number after ``FCD'' in the figure is the setting of $\Z$ in Algorithm \ref{ALG:alg2} for \scd.
Results show that the stage-wise case ($\Z=1$) of \scd is the fastest one, as
expected.
When we set $ \Z \geq 2 $,
the objective value of the optimization
\eqref{eq:mcgcon} of our method converges much faster than LBFGS-B.
Thus setting of $\Z=2$
is sufficient to achieve a very  accurate solution, and at the same
time
has faster convergence and less running time than LBFGS-B.

\section{Conclusion}

    In this work, we have presented a novel multi-class boosting method.
    Based on the dual problem, boosting is implemented using
    the column generation technique.
    Different from most existing multi-class boosting, we train a weak
    learner set for each class, which results in much faster
    convergence.

    A wide range of experiments on a few different datasets
    demonstrate that the proposed multi-class boosting achieves
    competitive test accuracy compared with other existing multi-class
    boosting. Yet it converges much faster and due to the proposed efficient
    coordinate descent method, the training of our method is much
    faster than the counterpart of \mb in \cite{Shen2011Direct}.

\noindent {\bf Acknowledgement}.
This work was supported by ARC grants LP120200485 and FT120100969.

\bibliographystyle{splncs}
\bibliography{CSRef}

\end{document}